\documentclass[11pt,draftclsnofoot,onecolumn]{IEEEtran}

\usepackage{graphicx}
\usepackage{epsfig}
\usepackage{amsmath}
\usepackage{bbm}
\usepackage{amssymb}
\usepackage{wrapfig}
\usepackage{times,color}
\usepackage{cite,footnote,xspace,syntonly,algorithm,algorithmic,bm}
\usepackage[caption=false,font=footnotesize]{subfig}

\usepackage{amsfonts}

\input{mysymbol.sty}
\newcommand{\calN}{{\cal N}}

\newcommand{\calP}{{\cal P}}

\def \cN {\mathcal{N}}
\def \cL {\mathcal{L}}
\def \cF {\mathcal{F}}

\def \cP {\mathcal{P}}

\def \bY {{\bf Y}}
\def \bR {{\bf R}}

\def \bX {{\bf X}}
\def \bA {{\bf A}}
\def \bV {{\bf V}}
\def \bW {{\bf W}}

\def \bG {{\bf G}}
\def \bI {{\bf I}}
\def \bB {{\bf B}}
\def \bM {{\bf M}}
\def \bU {{\bf U}}
\def \bV {{\bf V}}

\def \bD {{\bf D}}

\def \bW {{\bf W}}
\def \bH {{\bf H}}
\def \bL {{\bf L}}
\def \bQ {{\bf Q}}
\def \bC {{\bf C}}

\def \bOmega {{\bf \Omega}}
\def \bw {{\bf w}}

\def \bv {{\bf v}}
\def \bx {{\bf x}}

\def \ba {{\bf a}}
\def \bb {{\bf b}}
\def \bc {{\bf c}}

\def \bv {{\bf v}}
\def \by {{\bf y}}
\def \bq {{\bf q}}
\def \bl {{\bf l}}
\def \bs {{\bf s}}

\def \bo {{\bf o}}
\def \bxi {{\boldsymbol \gamma}}
\def \balpha {{\boldsymbol \alpha}}
\def \bbeta {{\boldsymbol \beta}}

\def \tr {\text{tr}}

\long\def\symbolfootnote[#1]#2{\begingroup
\def\thefootnote{\fnsymbol{footnote}}
\footnote[#1]{#2}\endgroup} \psfull

\begin{document}
%--------------------------------------------The First Page---------------------------------------------------------------
\title{\huge Subspace Learning and Imputation for\\ 
Streaming Big Data Matrices and Tensors$^\dag$}

\author{{\it Morteza Mardani, \textit{Student Member}, \textit{IEEE}, Gonzalo~Mateos, \textit{Member}, \textit{IEEE}, \\
and Georgios~B.~Giannakis, \textit{Fellow}, \textit{IEEE}$^\ast$}}

\markboth{IEEE TRANSACTIONS ON SIGNAL PROCESSING (SUBMITTED)}
\maketitle \maketitle \symbolfootnote[0]{$\dag$ Work in this paper
was supported by the MURI Grant No. AFOSR FA9550-10-1-0567. Part of the results in this paper were 
presented at the~{\it 38th IEEE International Conference on Acoustics, Speech, and Signal 
Processing}, Vancouver, Canada, May 2013; and were submitted to the 
{\it 8th IEEE Sensor Array and Multichannel Signal Processing Workshop}, A Coru\~{n}a, Spain, June 2014. } \symbolfootnote[0]{$\ast$ The 
authors are with the Dept.
of ECE and the Digital Technology Center, University of
Minnesota, 200 Union Street SE, Minneapolis, MN 55455. Tel/fax:
(612)626-7781/625-4583; Emails:
\texttt{\{morteza,mate0058,georgios\}@umn.edu}}

\vspace*{-75pt}
\begin{center}
\small{\bf Submitted: }\today\\
\end{center}
%\vspace*{10pt}

% % % % % % % % % % % % % % % % % % % % % % % % % % % % % % % % % % % % % % % %
%                         Abstract                                            %
% % % % % % % % % % % % % % % % % % % % % % % % % % % % % % % % % % % % % % % %

\thispagestyle{empty}\addtocounter{page}{-1}
\begin{abstract}
Extracting latent low-dimensional structure from high-dimensional data is of paramount 
importance in timely inference tasks encountered with `Big Data' analytics. However, 
increasingly noisy, heterogeneous, and incomplete datasets as well as the need for 
{\em real-time} processing of streaming data pose major challenges to this end. In this 
context, the present paper permeates benefits from rank minimization to scalable imputation of missing data,
via tracking low-dimensional subspaces and unraveling latent (possibly multi-way) structure 
from \emph{incomplete streaming} data. For low-rank matrix data, a subspace estimator is proposed based on an 
exponentially-weighted least-squares criterion regularized with the nuclear norm. 
After recasting the non-separable nuclear norm into a form amenable to online optimization, 
real-time algorithms with complementary strengths are developed and their convergence is established under simplifying 
technical assumptions. In a stationary setting, the asymptotic estimates 
obtained offer the well-documented performance guarantees of the 
{\em batch} nuclear-norm regularized estimator. Under the same
unifying framework, a novel online (adaptive) algorithm is developed to obtain
multi-way decompositions of \emph{low-rank tensors} with
missing entries, and perform imputation as a byproduct. Simulated 
tests with both synthetic as well as real Internet and cardiac
magnetic resonance imagery (MRI) data confirm the efficacy of the proposed
algorithms, and their superior performance relative to state-of-the-art alternatives.
\end{abstract}
\vspace*{-5pt}
%\newpage
\begin{center} \bfseries EDICS Category: SSP-SPRS, SAM-TNSR, OTH-BGDT. \end{center}
%
% for peerreview papers, inserts a page break and creates the second title.
% Will be ignored for other modes.
%\IEEEpeerreviewmaketitle

% % % % % % % % % % % % % % % % % % % % % % % % % % % % % % % % % % % % % % % %
%                         Section I                                           %
% % % % % % % % % % % % % % % % % % % % % % % % % % % % % % % % % % % % % % % %

\section{Introduction}\label{sec:intro}
Nowadays ubiquitous e-commerce sites, the Web, and Internet-friendly portable 
devices generate massive volumes of data. The overwhelming consensus is that 
tremendous economic growth and improvement in quality of life can be effected by 
harnessing the potential benefits of analyzing
this large volume of data. As a result,
the problem of extracting the most informative, yet low-dimensional structure from 
high-dimensional datasets is of paramount importance \cite{elements_of_statistics}. 
The sheer volume of data and the fact that oftentimes
observations are acquired sequentially in time, motivate updating previously 
obtained `analytics' rather than re-computing new ones from scratch each time a new 
datum becomes available~\cite{kostas_spmag_2014,mairalonlinelearning}. 
In addition, due to the disparate origins of the data, subsampling for faster data acquisition, or even
privacy constraints, the datasets are often incomplete~\cite{candes_moisy_mc,kolda_completion}. 

In this context, consider streaming data comprising incomplete and noisy observations
of the signal of interest $\bx_t\in \mathbb{R}^{P}$ at time $t=1,2,\ldots$. Depending on the application, these acquired vectors could e.g., correspond to (vectorized) images, link
traffic measurements collected across physical links of a computer network, or,
movie ratings provided by Netflix users. Suppose that 
the signal sequence $\{\bx_t\}_{t=1}^{\infty}$ lives in a \textit{low-dimensional} ($\ll P$) 
linear subspace $\cL_t$ of $\mathbb{R}^P$. Given the incomplete observations that are acquired sequentially in time, 
this paper deals first with (adaptive) online estimation of 
$\cL_t$, and reconstruction of the signal $\bx_t$ as a byproduct. 
This problem can be equivalently viewed as low-rank matrix
completion with noise~\cite{candes_moisy_mc}, solved online over 
$t$ indexing the columns of relevant matrices, e.g., $\bX_t:=[\bx_1,\ldots, \bx_t]$.

Modern datasets are oftentimes indexed
by three or more variables giving rise to a \emph{tensor}, that is a data cube or a mutli-way array, in
general~\cite{kolda_tutorial}. It is not uncommon 
that one of these variables indexes time~\cite{nion_online_tensor}, and that sizable portions of the 
data are missing~\cite{juan_tensor_tsp_2013,kolda_completion,GRY11,tensor_completion_visualdata_liu13,tensor_vs_matrix_completion_july2011}. Various data analytic tasks 
for network traffic, social networking, or medical data analysis aim at capturing
underlying latent structure, which calls for high-order tensor factorizations even 
in the presence of missing data~\cite{juan_tensor_tsp_2013,kolda_completion,tensor_completion_visualdata_liu13}. 
It is in principle possible to unfold the 
given tensor into a matrix and resort to either batch~\cite{RFP07,GRY11}, or, online matrix 
completion algorithms as the ones developed in the first part of this 
paper; see also~\cite{jstsp_anomalography_2012,onlinetracking_bolzano10,petrels_2013}. However,
tensor models preserve the multi-way nature of the data and extract the 
underlying factors in each mode (dimension) of a higher-order array.  
Accordingly, the present paper also contributes towards fulfilling a pressing need in terms 
of analyzing streaming and incomplete multi-way data; namely, \emph{low-complexity, real-time algorithms} 
capable of unraveling latent structures through parsimonious (e.g., low-rank) decompositions,
such as the parallel factor analysis (PARAFAC) model; see e.g.~\cite{kolda_tutorial} for a comprehensive
tutorial treatment on tensor decompositions, algorithms, and applications.

\noindent\textbf{Relation to prior work.}  Subspace tracking has a long history in signal
processing. An early noteworthy representative is the projection approximation subspace tracking (PAST) 
algorithm~\cite{yang95}; see also~\cite{yang_mos_88}. Recently, an algorithm (termed GROUSE) for tracking 
subspaces from incomplete observations
was put forth in~\cite{onlinetracking_bolzano10}, based on incremental gradient 
descent iterations on the Grassmannian manifold of subspaces. Recent analysis has shown that GROUSE can
converge locally at an expected linear rate~\cite{balzano_grouse_local_2013},
and that it is tightly related to the incremental SVD algorithm~\cite{balzano_incrementalsvd_2013}. 
PETRELS is a  second-order recursive least-squares (RLS)-type algorithm, that
extends the seminal PAST iterations to handle missing data~\cite{petrels_2013}. As noted in~\cite{SET-Dai-tsp2011}, the 
performance of GROUSE is limited by the existence of barriers in the search path on the Grassmanian, 
which may lead to GROUSE iterations being trapped at local minima; see also~\cite{petrels_2013}. 
Lack of regularization in PETRELS can also lead to unstable (even divergent) behaviors, 
especially when the amount of missing data is large. Accordingly, the convergence
results for PETRELS are confined to the full-data setting where the algorithm boils
down to PAST~\cite{petrels_2013}. Relative to all aforementioned works, 
the algorithmic framework of this paper permeates benefits from 
\emph{rank minimization} to low-dimensional subspace tracking 
and missing data imputation (Section \ref{sec:onlineMC}),
offers provable convergence and theoretical performance guarantees in a stationary setting 
(Section \ref{sec:perf_guarantees}),
and is flexible to accommodate tensor streaming data models as well (Section \ref{sec:tensor_completion}). 
While algorithms to impute incomplete 
tensors have been recently proposed in e.g.,~\cite{GRY11,juan_tensor_tsp_2013,kolda_completion,tensor_completion_visualdata_liu13}, 
all existing approaches rely on batch processing.

\noindent\textbf{Contributions.} Leveraging the low dimensionality of the underlying
subspace $\cL_t$, an estimator is proposed based on an exponentially-weighted
least-squares (EWLS) criterion regularized with the nuclear norm of $\bX_t$. For a related
data model, similar algorithmic construction ideas were put forth in our precursor paper~\cite{jstsp_anomalography_2012}, which dealt with real-time identification of network
traffic anomalies. Here instead, the focus is on subspace tracking from incomplete measurements, and online matrix
completion. Upon recasting the non-separable nuclear norm into a form 
amenable to online optimization as in~\cite{jstsp_anomalography_2012}, real-time subspace 
tracking algorithms with complementary strengths are developed
in Section \ref{sec:onlineMC}, and their convergence is established under simplifying technical 
assumptions.
For stationary data and under mild assumptions, the proposed online 
algorithms provably attain the global optimum of the batch nuclear-norm regularized 
problem (Section \ref{subsec:optimality}), whose quantifiable performance has 
well-appreciated merits~\cite{candes_moisy_mc,CR08}.
This optimality result as well as the convergence of the (first-order) stochastic-gradient
subspace tracker established in Section \ref{subsec:cnvg_stoc_gradient}, markedly broaden and complement the 
convergence claims in~\cite{jstsp_anomalography_2012}.

The present paper develops for
the first time an \emph{online} algorithm for decomposing \emph{low-rank 
tensors with missing entries}; see also~\cite{nion_online_tensor}
for an adaptive algorithm to obtain PARAFAC decompositions with full data.
Accurately approximating a given incomplete tensor allows one
to \emph{impute} those missing entries as a byproduct, by simply reconstructing the data cube 
from the model factors (which for PARAFAC are unique under relatively mild assumptions~\cite{jk77laa,tBS02}). 
Leveraging stochastic gradient-descent iterations, a scalable, real-time algorithm is 
developed in Section~\ref{sec:tensor_completion} under the same rank-minimization 
framework utilized for the matrix case, which here entails minimizing an EWLS 
fitting error criterion regularized by separable Frobenius norms of the 
PARAFAC decomposition factors~\cite{juan_tensor_tsp_2013}.
The proposed online algorithms offer a viable approach 
to solving large-scale tensor decomposition (and completion) problems, even if the data is not
actually streamed but they are so massive that do not fit in the main memory. 

Simulated tests with synthetic as well as real Internet 
traffic data corroborate the effectiveness of the proposed algorithms for
traffic estimation and anomaly detection, and its superior performance relative to state-of-the-art 
alternatives (available only for the matrix case~\cite{onlinetracking_bolzano10,petrels_2013}). 
Additional tests with cardiac magnetic resonance
imagery (MRI) data confirm the efficacy of the proposed tensor algorithm in imputing
up to $75\%$ missing entries. Conclusions are drawn in Section \ref{sec:discussions}.

\noindent{\it Notation}: Bold uppercase (lowercase) letters will denote
matrices (column vectors), and calligraphic letters will be used for sets.
Operators $(\cdot)'$, $\rm{tr}(\cdot)$, $\mathbbm{E}[\cdot]$, $\sigma_{\max}(\cdot)$, $\odot$, and $\circ$ will denote transposition, matrix trace, statistical expectation, maximum singular value, Hadamard product, and outer product, respectively; $|\cdot|$ will be used for the cardinality of a set, and the magnitude of a scalar. The positive semidefinite matrix $\mathbf{M}$ will be denoted by $\bbM\succeq\mathbf{0}$. The $\ell_p$-norm of $\bx \in \mathbb{R}^n$ is $\|\bx\|_p:=(\sum_{i=1}^n |x_i|^p)^{1/p}$ for $p \geq 1$. For two matrices $\bM,\bU \in \mathbb{R}^{n \times p}$,
$\langle \bM, \bU \rangle := \rm{tr(\bM' \bU)}$ denotes their trace inner 
product, and $\|\bM\|_F:=\sqrt{\tr(\bM\bM')}$ is the Frobenious norm. The $n \times n$ identity matrix will be represented by $\bI_n$, while $\mathbf{0}_{n}$ will stand for the $n \times 1$ vector of all zeros, $\mathbf{0}_{n \times p}:=\mathbf{0}_{n} \mathbf{0}'_{p}$, and $[n]:=\{1,2,\ldots,n\}$.

\section{Preliminaries and Problem Statement}
\label{sec:prob_satement}
Consider a sequence of high-dimensional data vectors, which are corrupted with additive noise
and some of their entries may be missing. At time $t$, the incomplete 
streaming observations are modeled as 
\begin{align}
\cP_{\omega_t}(\by_t) = \cP_{\omega_t}(\bx_t + \bv_t),\quad t =1,2,\ldots
\label{eq:data_model}
\end{align}
where $\bx_t\in \mathbb{R}^{P}$ is the signal of interest, and $\bv_t$ stands for the noise. 
The set $\omega_t \subset \{1,2,\ldots,P\}$ contains the indices of available observations, while the corresponding sampling operator $\cP_{\omega_t}(\cdot)$ sets the entries of its 
vector argument not in $\omega_t$ to zero, and keeps the rest unchanged; note that $\cP_{\omega_t}(\by_t) \in \mathbb{R}^P$. Suppose that 
the sequence $\{\bx_t\}_{t=1}^\infty$ lives in a \textit{low-dimensional} ($\ll P$) 
linear subspace $\cL_t$, which is allowed to change slowly over time. Given the incomplete observations $\{\cP_{\omega_{\tau}}(\by_{\tau})\}_{\tau=1}^t$, the first part of this paper deals with online (adaptive) estimation of 
$\cL_t$, and reconstruction of $\bx_t$ as a byproduct. The reconstruction here involves imputing the missing elements, and denoising the observed ones. 

\subsection{Challenges facing large-scale nuclear norm minimization}
\label{subsec:nuclear_norm}

\begin{figure}[t]
\centering
\includegraphics[scale=0.7]{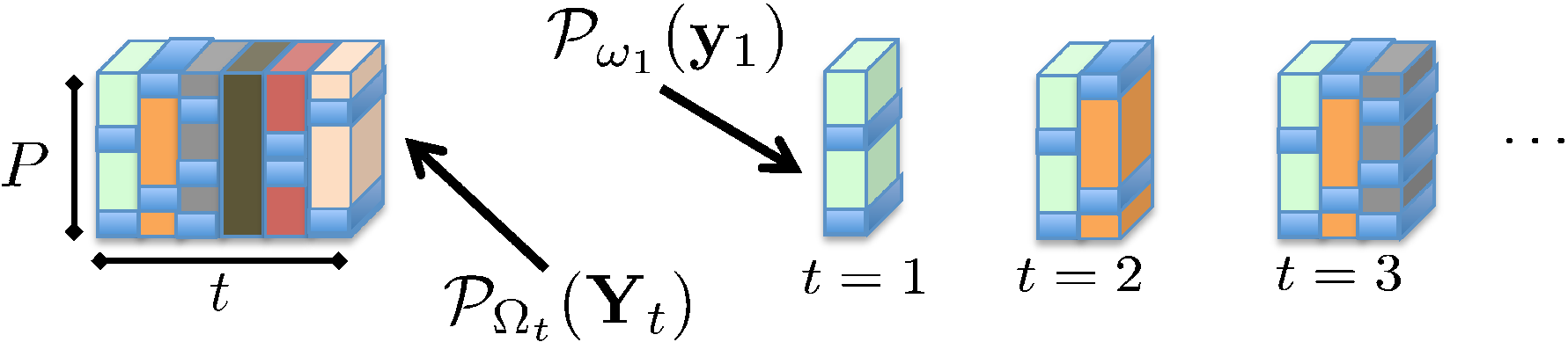}
\caption{Matrix data with missing entries. (Left) Batch data $\calP_{\Omega_t}(\bY_t)$ available at time $t$. (Right) Streaming data, where vectors $\calP_{\omega_t}(\by_t)$ become available for $t=1,2,\ldots$.}
 \label{fig:Fig_0}
\end{figure}

%\begin{figure}[t]
%\centering
%\epsfig{file=./figures/matrix_batch_stream_h.eps,width=0.6
%\linewidth, height=2.5 in }\caption{Matrix data with missing entries. (Left) Batch data $\calP_{\Omega_t}(\bY_t)$ available at time $t$. (Right) Streaming data, where vectors $\calP_{\omega_t}(\by_t)$ become available for $t=1,2,\ldots$.}
% \label{fig:Fig_0}
%\end{figure}

Collect the indices of available observations up to time $t$ in the set 
$\Omega_t:=\cup_{\tau=1}^t \omega_\tau$, and the actual batch of observations in 
the matrix $\cP_{\Omega_t}(\bY_t):=[\cP_{\omega_1}(\by_1),\ldots,\cP_{\omega_t}(\by_t)] \in \mathbb{R}^{P \times t}$;
see also Fig. \ref{fig:Fig_0}. 
Likewise, introduce matrix $\bX_t$ containing the signal of interest. Since $\bx_t$ lies in a low-dimensional 
subspace, $\bX_t$ is (approximately) a \emph{low-rank} matrix. A natural estimator leveraging the low rank property of $\bX_t$ 
attempts to fit the incomplete data $\cP_{\Omega_t}(\bY_t)$ to $\bX_t$ in the least-squares~(LS) sense, 
as well as minimize the rank of $\bX_t$. Unfortunately, albeit natural the rank criterion is in general NP-hard to 
optimize~\cite{RFP07}. This motivates solving for~\cite{candes_moisy_mc}
\begin{align}
\text{(P1)}~~~~\hat{\bX}_t:=\arg\min_{\bX} \left\{\frac{1}{2} \|\cP_{\Omega_t}(\bY_t-\bX) \|_F^2
+ \lambda_t \|\bX\|_{\ast} \right\}\nonumber% \label{eq:batch_est_nuclear}
\end{align}
where the nuclear norm $\|\bX_t\|_*:=\sum_{k}\sigma_k(\bX_t)$ ($\sigma_k$ is the 
$k$-th singular value) is adopted as a convex surrogate to $\textrm{rank}(\bX_t)$~\cite{fazel_phdthesis}, and $\lambda_t$ is a (possibly time-varying) rank-controlling parameter. 
Scalable imputation algorithms for 
streaming observations should effectively overcome the following challenges: (c1) the 
problem size can easily become quite large, since the number of optimization variables~$P t$ grows with time; (c2) 
existing batch iterative solvers for (P1) typically rely on costly SVD computations per iteration; 
see e.g.,~\cite{CR08}; and (c3) (columnwise) nonseparability 
of the nuclear-norm challenges online processing when new columns $\{\cP_{\omega_t}(\by_t)\}$ 
arrive sequentially in time. In the following subsection, the `Big Data' challenges (c1)-(c3) are 
dealt with to arrive at an efficient online algorithm in Section~\ref{sec:onlineMC}.

\subsection{A separable low-rank regularization}
\label{subsec:separable}
To limit the computational complexity and memory
storage requirements of the algorithm sought, it is henceforth assumed that the dimensionality 
of the underlying time-varying subspace $\mathcal{L}_t$ is bounded by a known quantity $\rho$. Accordingly, 
it is natural to require $\textrm{rank}(\hat{\bX}_t)\leq \rho$. 
As argued later in Remark \ref{rem:remark_1}, the smaller the value of $\rho$, the more efficient 
the algorithm becomes. Because 
$\textrm{rank}(\hat{\bX}_t)\leq \rho$ one can factorize 
the matrix decision variable as $\bX=\bL\bQ'$, where $\bL$ and $\bQ$ are $P \times \rho$ and $t \times \rho$ 
matrices, respectively. Such a bilinear decomposition suggests $\mathcal{L}_t$ is spanned by the columns of $\bL$, 
while the rows of $\bQ$ are 
the projections of $\{\bx_t\}$ onto $\mathcal{L}_t$.

To address (c1) and (c2) [along with (c3) as it will become clear in Section \ref{sec:onlineMC}], 
consider the following alternative characterization of the nuclear norm~\cite{srebro_2005}
\begin{equation}\label{eq:nuc_nrom_def}
\|\bX\|_*:=\min_{\{\bL,\bQ\}}~~~ \frac{1}{2}\left\{\|\bL\|_F^2+\|\bQ\|_F^2 \right\},\quad
\text{s. to}~~~ \bX=\bL\bQ'. 
\end{equation}
The optimization \eqref{eq:nuc_nrom_def} is over all possible bilinear 
factorizations of $\bX$, so that the number of columns $\rho$ of $\bL$ 
and $\mathbf{Q}$ is also a variable. Leveraging~\eqref{eq:nuc_nrom_def}, the following nonconvex reformulation
of (P1) provides an important first step towards obtaining an online algorithm:
\begin{align}
\text{(P2)}~~~~~\min_{\{\bL,\bQ\}}& \frac{1}{2}\|\cP_{\Omega_t}(\bY_t - \bL\bQ')\|_{F}^{2} + 
\frac{\lambda_t}{2}\left\{\|\bL\|_F^2 + \|\bQ\|_F^2 \right\}. \nonumber
\end{align}
The number of variables is reduced from $P t$ in (P1) 
to $\rho(P+t)$ in (P2), which can be significant when $\rho$ is small, and both $P$ and $t$ are large.
Most importantly, it follows that adopting the separable (across the time-indexed columns of $\bQ$) Frobenius-norm 
regularization in (P2) comes with no loss of optimality relative to (P1), provided 
$\rho\geq\textrm{rank}({\hat\bX}_t)$. 

By finding the global minimum of (P2), 
one can recover the optimal solution of (P1). However, since (P2) is nonconvex,
it may have stationary points which need not be globally optimum. Interestingly, results in~\cite{tsp_rankminimization_2012,burer2005local} offer a global optimality certificate for 
stationary points of (P2). Specifically, if $\{\bar{\bL}_t,\bar{\bQ}_t\}$ is a stationary point of (P2)
(obtained with any practical solver) satisfying the qualification inequality $\sigma_{\max}[\cP_{\Omega_t}(\bY_t-\bar{\bL}_t\bar{\bQ}'_t)] \leq \lambda_t$, then ${\hat{\bX}}_t:=\bar{\bL}_t\bar{\bQ}'_t$ is the globally optimal solution of (P1)~\cite{tsp_rankminimization_2012,burer2005local}.

%\begin{remark}(Optimality)\label{rem:remark_1}
%Verifying the qualification condition indeed needs knowing the converged solutions $\{\bar{\bL}_t,\bar{\bQ}_t\}$. However, empirical observations evidence that if the parameters $\lambda$ and $\rho$ are chosen large enough,; not necessarily too large to make $\bar{\bL}_t=\mathbf{0}_{P\times \rho}$ and~$\bar{\bQ}_t=\mathbf{0}_{t\times \rho}$, then $\hat{\bX}_t=\bar{\bL}_t\bar{\bQ}_t^{'}$ qualify global optimality; see section~\ref{sec:sims}. Analytic proof of this interesting observation seems tricky, and will be pursued as a future research. 
%\end{remark}

% % % % % % % % % % % % % % % % % % % % % % % % % % % % % % % % % % % % % % % %
%                         Section III                                        %
% % % % % % % % % % % % % % % % % % % % % % % % % % % % % % % % % % % % % % % %

\section{Online Rank Minization for Matrix Imputation}
\label{sec:onlineMC}
In `Big Data' applications the collection of massive amounts of data far outweigh 
the ability of modern computers to store and analyze them as a batch. In addition, 
in practice (possibly incomplete) observations are acquired sequentially in time 
which motivates updating previously obtained estimates rather than re-computing 
new ones from scratch each time a new datum becomes available. As stated in 
Section~\ref{sec:prob_satement}, the goal is to recursively track 
the low-dimensional subspace $\cL_t$, and subsequently estimate $\hat{\bx}_t$ per 
time $t$ from historical observations $\{\cP_{\omega_\tau}(\by_\tau)\}_{\tau=1}^t$, 
naturally placing more importance on recent measurements. To this end, one possible 
adaptive counterpart to~(P2) is the exponentially-weighted LS (EWLS) estimator found by 
minimizing the empirical cost
\begin{align}
{\rm (P3)}~~~ \min_{\{\bL,\bQ\}} \sum_{\tau=1}^t \theta^{t-\tau} \left[ \frac{1}{2}
\|\cP_{\omega_\tau}(\by_\tau-\bL\bq_\tau) \|_2^2 + \frac{\bar{\lambda}_t}{2} \|\bL\|_F^2 +  \frac{\lambda_t}{2} \|\bq_\tau\|_2^2
 \right]\nonumber %\label{eq:adaptive_v1_est_lq}
\end{align}
where $\bQ:=[\bq_1,\ldots,\bq_t]$, $\bar{\lambda}_t:=\lambda_t/\sum_{\tau=1}^t \theta^{t-\tau}$, and $ 0< \theta \leq 1$ 
is the so-termed forgetting factor. When $\theta<1$, data in the distant past are exponentially 
downweighted, which facilitates tracking in nonstationary environments. In the case of 
infinite memory $(\theta=1)$, the formulation~(P3) coincides 
with the batch estimator (P2). This is the reason for the time-varying factor $\bar{\lambda}_t$ weighting $\|\bL\|_F^2$.

We first introduced the basic idea of performing online rank-minimization leveraging the separable nuclear-norm regularization 
\eqref{eq:nuc_nrom_def} in~\cite{jstsp_anomalography_2012} (and its
conference precursor), in the context of unveiling network traffic anomalies. Since then, the approach
has gained popularity in real-time non-negative matrix factorization
for singing voice separation from its music accompaniment~\cite{pablo_ismir2012}, 
and online robust PCA~\cite{orpca_nips}, too name a few examples. Instead, the novelty here is on subspace tracking from 
incomplete measurements, as well as online low-rank matrix and tensor completion.

\subsection{Alternating recursive LS for subspace tracking from incomplete data}
\label{subsec:tracking_missing}
Towards deriving a real-time, computationally efficient, and recursive solver of~(P3),
an alternating-minimization (AM) method is adopted in which iterations coincide with the 
time-scale $t$ of data acquisition. A justification in terms of minimizing a suitable approximate cost 
function is discussed in detail in Section \ref{subsec:cnvg_secondorder_alg}. Per time instant $t$, a new datum 
$\{\cP_{\omega_t}(\by_t)\}$ is drawn and $\bq_t$ is estimated via
\begin{align}
\bq[t] = \arg\min_{\bq} \left[ \frac{1}{2}
\|\cP_{\omega_t}(\by_t-\bL[t-1]\bq)\|_2^2 +
\frac{\lambda_t}{2} \|\bq\|_2^2\right] \label{eq:rec_est_q}
\end{align}
which is an $\ell_2$-norm regularized LS (ridge-regression) problem. It admits the closed-form solution
\begin{align}
\bq[t] &= \left( \lambda_t \bI_{\rho} +  \bL'[t-1] \mathbf{\Omega}_t \bL[t-1]
\right)^{-1}
\bL'[t-1]\cP_{\omega_t}(\by_t)  \label{eq:q_ls}
\end{align}
where diagonal matrix $\mathbf{\Omega}_t \in \{0,1\}^{P \times P}$ is such that $[\mathbf{\Omega}_t]_{p,p}=1$ if $p\in\omega_t$, and is zero elsewhere. In the second step of the AM scheme, the updated subspace matrix $\bL[t]$ is obtained by minimizing~(P3) with respect to $\bL$, while the optimization variables $\{\bq_{\tau}\}_{\tau=1}^t$ are fixed and take the values $\{\bq[\tau]\}_{\tau=1}^t$, namely
\begin{align}
\bL[t] = \arg\min_{\bL} \left[\frac{\lambda_t}{2} \|\bL\|_F^2 + \sum_{\tau=1}^t  \theta^{t-\tau}
\frac{1}{2} \|\cP_{\omega_\tau}(\by_\tau -
\bL\bq[\tau])\|_2^2\right]. \label{eq:rec_est_L}
\end{align}
Notice that \eqref{eq:rec_est_L} decouples over the rows of $\bL$ which are obtained in parallel via 
\begin{align}
\bl_{p}[t] = &\arg\min_{\bl} \left[ \frac{\lambda_t}{2} 
\|\bl\|_2^2+\sum_{\tau=1}^t \theta^{t-\tau}
\omega_{p,\tau}(y_{p,\tau} - \bl'\bq[\tau])^2\right],  
\label{eq:rec_est_li}
\end{align}
for $p=1,\ldots,P$, where $\omega_{p,\tau}$ denotes the $p$-th diagonal entry of $\bm{\Omega}_{\tau}$. For $\theta=1$ and fixed $\lambda_t=\lambda,~\forall t$, subproblems \eqref{eq:rec_est_li} can be efficiently solved using recursive LS  (RLS)~\cite{Solo_Adaptive_Book}. Upon defining $\bs_p[t]:=\sum_{\tau=1}^t\theta^{t-\tau}\omega_{p,\tau}y_{p,\tau}\bq[\tau]$, $\bH_p[t]:=\sum_{\tau=1}^t\theta^{t-\tau}\omega_{p,\tau}\bq[\tau]\bq'[\tau]+\lambda_t \bI_\rho$, and $\bM_{p}[t]:=\bH_p^{-1}[t]$, one updates 
\begin{align}
\bs_{p}[t]{}={} & \bs_{p}[t-1] +\omega_{p,t}y_{p,t} \bq[t]\nonumber\\
\bM_{p}[t] {}={} & \bM_{p}[t-1] 
-\omega_{p,t}\frac{\bM_p[t-1]\bq[t]\bq'[t]\bM_p[t-1]}{1+\bq'[t]\bM_p[t-1]\bq[t]}\nonumber
\end{align}
and forms $\bl_{p}[t]=\bM_{p}[t]\bs_{p}[t]$, for $p=1,\ldots,P$.

However, for $0< \theta<1$ the regularization term $(\lambda_t/2) \|\bl\|_2^2$ in 
\eqref{eq:rec_est_li} makes it impossible to express $\bH_p[t]$ in terms of $\bH_p[t-1]$ 
plus a rank-one correction. Hence, one cannot resort to the matrix inversion lemma and 
update $\bM_{p}[t]$ with quadratic complexity only. Based on direct inversion of each 
$\bH_p[t]$, the alternating recursive LS algorithm for subspace tracking from 
incomplete data is tabulated under Algorithm~\ref{tab:table_1}.

%The per iteration cost of the $L$ inversions (each $\mathcal{O}(\rho^3)$, which could be further reduced if one leverages also the symmetry of $\bH_l[t]$) is affordable for moderate $L$, because $\rho$ is small when estimating low-rank traffic matrices. 

\begin{algorithm}[t]
\caption{: Alternating LS for subspace tracking from incomplete observations} \small{
\begin{algorithmic}
	\STATE \textbf{input} 
	$\{\cP_{\omega_{\tau}}(\by_{\tau}),\omega_{\tau}\}_{{\tau}=1}^{\infty}$, $\{\lambda_{\tau}\}_{\tau=1}^{\infty}$, and $\theta$.
    \STATE \textbf{initialize} $\bG_{p}[0]=\mathbf{0}_{\rho\times \rho}$,
    $\bs_{p}[0]=\mathbf{0}_{\rho},~p=1,...,P$, and $\bL[0]$ at
    random.
    \FOR {$t=1,2$,$\ldots$}
                \STATE $\bD[t] = \left(\lambda_t \bI_{\rho} + \bL'[t-1]
                \mathbf{\Omega}_t \bL[t-1]\right)^{-1} \bL'[t-1]$.
%                \STATE $\bF[t]= \left[\mathbf{\Omega}_t -
%                \mathbf{\Omega}_t\bL[t-1]\bD[t]\mathbf{\Omega}_t, \sqrt{\lambda_{\ast}}
%                \mathbf{\Omega}_t \bD'[t]\right]'$.
%                \STATE $\ba[t] = \arg\min_{\ba} \left[\frac{1}{2}
%                \|\bF[t](\by_t- \bR_t\ba)\|^2 + \lambda_1 \|\ba\|_1 \right]$.
                \STATE $\bq[t] = \bD[t] \cP_{\omega_t}(\by_t)$.
                \STATE $\bG_{p}[t] = \theta \bG_{p}[t-1] + \omega_{p,t}
                \bq[t]\bq[t]', \quad p=1,\ldots,P$.
                \STATE $\bs_{p}[t]=\theta \bs_{p}[t-1] +
                \omega_{p,t} y_{p,t} \bq[t], \quad
                p=1,\ldots,P$.
                 \STATE $\bl_{p}[t] = \left(\bG_{p}[t] + \lambda_t
                 \bI_{\rho}\right)^{-1} \bs_{p}[t], \quad p=1,...,P$.
                \RETURN  $\hat{\bx}_t := \bL[t] \bq[t]$.
    \ENDFOR
\end{algorithmic}}
\label{tab:table_1}
\end{algorithm}

\begin{remark}[Computational cost]\label{rem:remark_1} Careful inspection of Algorithm~\ref{tab:table_1} 
reveals that the main computational burden stems from $\rho\times\rho$ inversions to update the subspace 
matrix $\bL[t]$. The per iteration complexity for performing the inversions is $ \mathcal{O}(P\rho^3)$ 
(which could be further reduced if one leverages also the symmetry of $\bG_{p}[t]$), while the cost for 
the rest of operations including multiplication and additions is $ \mathcal{O}(P\rho^2)$. The overall 
cost of the algorithm per iteration can thus be safely estimated as $\mathcal{O}(P\rho^3)$, which can 
be affordable since $\rho$ is typically small (cf. the low rank assumption). In addition, for the 
infinite memory case $\theta=1$ where the RLS update is employed, the overall cost is further reduced 
to $\mathcal{O}(|\omega_t|\rho^2)$. 
\end{remark}

\begin{remark}[Tuning ${\lambda_t}$] \label{rem:remark_2} 
To tune $\lambda_t$  one can resort to the heuristic rules proposed in~\cite{candes_moisy_mc}, 
which apply under the following assumptions: i) $v_{p,t} \sim \cN(0,\sigma^2)$; ii) elements of 
$\Omega_t$ are independently sampled with probability $\pi$; and, iii) $P$ and $t$ are large enough. 
Accordingly, one can pick $\lambda_t=\big(\sqrt{P}+\sqrt{t_{e}}\big)\sqrt{\pi}\sigma$, where 
$t_{e}:=\sum_{\tau=1}^t \theta^{t-\tau}$ is the effective time window. Note that $\lambda_t$ naturally
increases with time when $\theta=1$, whereas for $\theta<1$ a fixed value $\lambda_t=\lambda$ is well
justified since the data window is effectively finite.
\end{remark}

\subsection{Low-complexity stochastic-gradient subspace updates}
\label{subsec:low_compx_sub_update}

Towards reducing Algorithm's \ref{tab:table_1} computational complexity in updating
the subspace $\bL[t]$, this section aims at developing lightweight algorithms 
which better suit the `Big Data' landscape. To this end, the basic AM
framework in Section~\ref{subsec:tracking_missing} will be retained,
and the update for $\bq[t]$ will be identical [cf. \eqref{eq:q_ls}]. However, 
instead of exactly solving an unconstrained quadratic program per iteration to 
obtain $\bL[t]$ [cf.~\eqref{eq:rec_est_L}], the subspace estimates will be 
obtained via stochastic-gradient descent (SGD) iterations.  
As will be shown later on, these updates can be traced to 
inexact solutions of a certain quadratic program different from \eqref{eq:rec_est_L}. 

For $\theta=1$, it is shown in Section \ref{subsec:cnvg_secondorder_alg} that Algorithm \ref{tab:table_1}'s  subspace estimate $\bL[t]$
is obtained by minimizing the empirical cost function $\hat{C}_t(\bL)=(1/t)\sum_{\tau=1}^t f_\tau(\bL)$, where
\begin{equation}\label{eq:f_t}
f_t(\bL):=\frac{1}{2}\|\cP_{{\omega}_t}(\bby_t-\bL\bbq[t])\|_2^2+
\frac{\lambda}{2t}\|\bL\|_F^2 +
\frac{\lambda}{2}
\|\bq[t]\|_2^2, \quad t=1,2,\ldots
\end{equation}
By the law of large numbers, if data $\{\cP_{\omega_t}(\by_t)\}_{t=1}^{\infty}$ are stationary, 
solving $\min_{\bL}\lim_{t\to\infty}\hat{C}_t(\bL)$ yields the desired minimizer of the \textit{expected} cost $\mathbb{E}[C_t(\bL)]$, where the expectation is taken with respect to the unknown probability 
distribution of the data. A standard approach to achieve this same goal -- typically with
reduced computational complexity -- is to drop the expectation (or the sample averaging
operator for that matter), and update the subspace via SGD; see e.g.,~\cite{Solo_Adaptive_Book}
\begin{align}\label{eq:stochastic_gradient}
\bL[t]{}=\bL[t-1]-(\mu[t])^{-1}\nabla f_t(\bL[t-1])
\end{align}
where $(\mu[t])^{-1}$ is the step size, and $\nabla f_t(\bL)=-\cP_{{\omega}_t}(\bby_t-\bL\bbq[t])\bbq'[t]+(\lambda/t)\bL$. 
The subspace update $\bL[t]$ is nothing but the minimizer of a second-order approximation $Q_{\mu[t],t}(\bL,\bL[t-1])$ of $f_t(\bL)$ around the previous subspace $\bL[t-1]$, where
\begin{align}
Q_{\mu,t}(\bL_1,\bL_2):=f_t(\bL_2)+\langle\bL_1-\bL_2,\nabla
f_t(\bL_2)\rangle +\frac{\mu}{2}\|\bL_1-\bL_2\|_f^2.
\end{align}
To tune the step size, the backtracking rule is adopted, whereby the non-increasing 
step size sequence $\{(\mu[t])^{-1}\}$ decreases geometrically at certain iterations 
to guarantee the quadratic function $Q_{\mu[t],t}(\bL,\bL[t-1])$ majorizes $f_t(\bL)$ 
at the new update $\bL[t]$. Other choices of the step size are discussed in 
Section~\ref{sec:perf_guarantees}. It is observed that different from Algorithm~\ref{tab:table_1}, 
no matrix inversions are involved in the update of the subspace $\bL[t]$. In the context of 
adaptive filtering, first-order SGD algorithms such as \eqref{eq:f_t} 
are known to converge slower than RLS. This is expected
since RLS can be shown to be an instance of Newton's (second-order) optimization 
method~\cite[Ch. 4]{Solo_Adaptive_Book}.

\begin{algorithm}[t]
\caption{: Online SGD for subspace tracking from incomplete observations} \small{
\begin{algorithmic}
	\STATE \textbf{input} $\{\cP_{\omega_{\tau}}(\by_{\tau}),\omega_{\tau}\}_{\tau=1}^\infty,$
	$\rho, \lambda,\eta>1$.
    \STATE \textbf{initialize} $\bL[0]$ at random, $\mu[0]>0$, $\tilde\bL[1]:=\bL[0]$, and $k[1]:=1$.
    \FOR {$t=1,2$,$\ldots$}
                \STATE $\bD[t] = \left(\lambda \bI_{\rho} + \bL'[t-1]
                \mathbf{\Omega}_t \bL[t-1]\right)^{-1} \bL'[t-1]$
                \STATE $\bq[t] = \bD[t] \cP_{\omega_t}(\by_t)$
        \STATE Find the smallest nonnegative integer $i[t]$ such that with
        $\bar{\mu}:=\eta^{i[t]}\mu[t-1]$
        $$f_t(\tilde\bL[t]-(1/\bar{\mu})\nabla f_t(\tilde\bL[t]))\leq
        Q_{\bar{\mu},t}(\tilde\bL[t]-(1/\bar{\mu})\nabla
        f_t(\tilde\bL[t]),\tilde\bL[t])$$
        holds, and set $\mu[t]=\eta^{i[t]}\mu[t-1].$
        \STATE $\bL[t]=\tilde\bL[t]-(1/\mu[t])\nabla f_t(\tilde\bL[t]).$
        \STATE $k[t+1]=\frac{1+\sqrt{1+4k^2[t]}}{2}.$
        \STATE
        $\tilde\bL[t+1]=\bL[t]+\left(\frac{k[t]-1}{k[t+1]}\right)(\bL[t]-\bL[t-
        1]).$
    \ENDFOR
    \RETURN $\hat{\bbx}[t]:=\bL[t]\bbq[t]$.
\end{algorithmic}}
\label{tab:table_2}
\end{algorithm}

Building on the increasingly popular \textit{accelerated} gradient methods for batch
smooth optimization~\cite{nesterov83,fista}, the idea here is to speed-up the 
learning rate of the estimated subspace \eqref{eq:stochastic_gradient}, 
without paying a penalty in terms of computational complexity per iteration.
The critical difference between standard gradient algorithms and the so-termed 
Nesterov's variant, is that the accelerated updates take the form $\bL[t]=\tilde\bL[t]-(\mu[t])^{-1}\nabla f_t(\tilde\bL[t])$, which relies on a judicious linear combination $\tilde\bL[t-1]$ of the previous pair of iterates $\{\bL[t-1],\bL[t-2]\}$. Specifically, the choice $\tilde\bL[t]=\bL[t-1]+\frac{k[t-1]-1}{k[t]}\left(\bL[t-1]-\bL[t-2]\right)$, where
$k[t]=\left[1+\sqrt{4k^2[t-1]+1}\right]/2$, has been shown to significantly
accelerate batch gradient algorithms resulting in convergence rate no worse than 
$\mathcal{O}(1/k^2)$; see e.g.,~\cite{fista} and references therein. Using this acceleration
technique in conjunction with a backtracking stepsize rule~\cite{Bers}, a fast
online SGD algorithm for imputing missing entries is tabulated
under Algorithm \ref{tab:table_2}. Clearly, a standard (non accelerated) SGD 
algorithm with backtracking step size rule is subsumed as a special
case, when $k[t]=1$, $t=1,2,\ldots$. In this case, complexity is 
$\mathcal{O}(|\omega_t|\rho^2)$ mainly due to update of $\bq_t$, while the 
accelerated algorithm incurs an additional cost $O(P\rho)$ for the subspace 
extrapolation step.

% % % % % % % % % % % % % % % % % % % % % % % % % % % % % % % % % % % % % % % %
%                         Section II                                          %
% % % % % % % % % % % % % % % % % % % % % % % % % % % % % % % % % % % % % % % %

\section{Performance Guarantees}
\label{sec:perf_guarantees}
This section studies the performance of the proposed first- and second-order online 
algorithms for the infinite memory special case; that is $\theta=1$. In the sequel,
 to make the analysis tractable the following assumptions are adopted:

\begin{description}
\item{(A1) Processes $\{\omega_t\}_{t=1}^\infty$ and 
$\{\cP_{\omega_t}(\by_t)\}_{t=1}^\infty$ are independent and identically distributed (i.i.d.);}

\item{(A2) Sequence $\{\cP_{\omega_t}(\by_t)\}_{t=1}^{\infty}$ is uniformly bounded; and}

\item{(A3) Iterates $\{\bL[t]\}_{t=1}^\infty$ lie in a compact set.}

\end{description}

To clearly delineate the scope of the analysis, it is worth commenting on (A1)--(A3) and the factors that influence their satisfaction. Regarding (A1), 
the acquired data is assumed statistically independent across time as it is customary 
when studying the stability and performance of online (adaptive) algorithms~\cite{Solo_Adaptive_Book}. 
While independence is required for tractability, (A1) may be grossly
violated because the observations $\{\cP_{\omega_t}(\by_t)\}$ are correlated across time 
(cf. the fact that $\{\bx_t\}$ lies in a low-dimensional subspace). Still, in accordance 
with the adaptive filtering folklore e.g.,~\cite{Solo_Adaptive_Book}, as $\theta \to 1$ or $(\mu[t])^{-1}\to 0$
the upshot of the analysis based on i.i.d. data extends accurately to the pragmatic setting 
whereby the observations are correlated. Uniform boundedness of $\cP_{\omega_t}(\by_t)$ 
[cf. (A2)] is natural in practice as it is imposed by the data acquisition process. The bounded 
subspace requirement in (A3) is a technical assumption that simplifies the analysis, and 
has been corroborated via extensive computer simulations.

\subsection{Convergence analysis of the second-order algorithm}
\label{subsec:cnvg_secondorder_alg}

Convergence of the iterates generated by Algorithm~\ref{tab:table_1} (with $\theta=1$) is established first. 
Upon defining
\begin{align}
g_t(\bL,\bq):=  \frac{1}{2}\|\cP_{\omega_t}(\by_t-\bL\bq)\|_2^2 +
\frac{\lambda_t}{2}\|\bq\|_2^2  \nonumber %\label{eq:rec_est_q}
\end{align}
in addition to $\ell_t(\bL):=\min_{\bq}g_t(\bL,\bq)$, Algorithm~\ref{tab:table_1} aims 
at minimizing the following \textit{average} cost function at time $t$
\begin{align}
C_t(\bL) := \frac{1}{t} \sum_{\tau=1}^t \ell_\tau(\bL) + 
\frac{\lambda_t}{2t}\|\bL\|_F^2. \label{eq:cost_target}
\end{align}
Normalization (by $t$) ensures that the cost function does not grow unbounded as time 
evolves. For any finite $t$, \eqref{eq:cost_target} is essentially identical to the batch
estimator in (P2) up to a scaling, which does not affect the value of the minimizer. Note 
that as time evolves, minimization of $C_t$
becomes increasingly complex computationally. Hence, at time $t$ the subspace estimate $\bL[t]$ 
is obtained by minimizing the \textit{approximate} cost function
\begin{align}
\hat{C}_t(\bL) = \frac{1}{t} \sum_{\tau=1}^t g_\tau(\bL,\bq[\tau]) +
\frac{\lambda_t}{2t} \|\bL\|_F^2 \label{eq:cost_apx}
\end{align}
in which $\bq[t]$ is obtained based on the prior subspace estimate $\bL[t-1]$ after solving 
$\bq[t]=\arg\min_{\bq} g_t(\bL[t-1],\bq)$ [cf. \eqref{eq:rec_est_q}]. Obtaining $\bq[t]$ this 
way resembles the projection approximation adopted in~\cite{yang95}. Since $\hat{C}_t(\bL)$ 
is a smooth convex quadratic function, the minimizer 
$\bL[t]=\arg\min_{\bL}\hat{C}_t(\bL)$ is the solution of the linear equation 
$\nabla \hat{C}_t(\bL[t]) = \mathbf{0}_{P\times \rho}$. 

So far, it is apparent that since $g_t(\bL,\bq[t]) \geq \min_{\bq} g_t(\bL,\bq)=\ell_t(\bL)$, 
the approximate cost function $\hat{C}_t(\bL[t])$ overestimates the target cost $C_t(\bL[t])$, 
for $t=1,2,\ldots$. However, it is not clear whether the subspace
iterates $\{\bL[t]\}_{t=1}^\infty$ converge, and most importantly, how well can they optimize 
the target cost function $C_t$. The good news is that $\hat{C}_t(\bL[t])$ asymptotically 
approaches $C_t(\bL[t])$, and the subspace iterates null $\nabla C_t(\bL[t])$ as well, both 
as $t\to\infty$. 
This result is summarized in the next proposition.

\begin{proposition}\label{prop:prop_2}
Under (A1)--(A3) and $\theta=1$ in Algorithm \ref{tab:table_1}, if $\lambda_t=\lambda~\forall t$ and 
$\lambda_{\min}[\nabla^2 \hat{C}_t(\bL)] \geq c$ for some $c>0$, then 
$\lim_{t \rightarrow \infty} \nabla C_t(\bL[t]) = \mathbf{0}_{P\times\rho}$ almost surely (a.s.), i.e., 
the subspace iterates $\{\bL[t]\}_{t=1}^\infty$ asymptotically fall into the stationary point set of 
the batch problem (P2).
\end{proposition}

It is worth noting that the pattern and the amount of misses, summarized in the sampling sets 
$\{\omega_t\}$, play a key role towards satisfying the Hessian's positive semi-definiteness condition. 
In fact, random misses are desirable since the Hessian $\nabla^2 \hat{C}_t(\bL)= \frac{\lambda}{t}
\bI_{P\rho} +\frac{1}{t}\sum_{\tau=1}^t (\bq[\tau] \bq'[\tau]) \otimes \mathbf{\Omega}_\tau $ is
 more likely to satisfy $\nabla^2 \hat{C}_t(\bL) \succeq c\bI_{P\rho}$, for some $c>0$. 

The proof of Proposition~\ref{prop:prop_2} is inspired by~\cite{mairalonlinelearning} which 
establishes convergence of an online dictionary learning algorithm using the theory of martingale 
sequences. Details can be found in our companion paper~\cite{jstsp_anomalography_2012}, and
in a nutshell the 
proof procedure proceeds in the following two main steps:

\noindent\textbf{(S1)} Establish that the approximate cost sequence $\{\hat{C}_t(\bL[t])\}$ asymptotically 
converges to the target cost sequence $\{C_t(\bL[t])\}$. To this end, it is first proved 
that $\{\hat{C}_t(\bL[t])\}_{t=1}^{\infty}$ is a quasi-martingale sequence, and hence 
convergent a.s. This relies on the fact that $g_t(\bL,\bq[t])$ is a {\it tight} upper bound 
approximation of $\ell_t(\bL)$ at the previous update $\bL[t-1]$, namely, 
$g_t(\bL,\bq[t]) \geq \ell_t(\bL),~\forall \bL \in \mathbb{R}^{P \times \rho}$,
and $g_t(\bL[t-1],\bq[t])=\ell_t(\bL[t-1])$. 

\noindent\textbf{(S2)} Under certain regularity assumptions on $g_t$, establish that 
convergence of the cost sequence $\{\hat{C}_t(\bL[t])-C_t(\bL[t])\} \rightarrow 0$ 
yields convergence of the gradients $\{\nabla\hat{C}_t(\bL[t])-\nabla C_t(\bL[t])\} \rightarrow 0$, 
which subsequently results in~$\lim_{t \rightarrow \infty} \nabla C_t(\bL[t])=\mathbf{0}_{P\times \rho}$. 

\subsection{Convergence analysis of the first-order algorithm}
\label{subsec:cnvg_stoc_gradient}

Convergence of the SGD iterates (without Nesterov's acceleration) is established here, by resorting 
to the proof techniques adopted for the second-order algorithm in 
Section~\ref{subsec:cnvg_secondorder_alg}. The basic idea is to judiciously derive an 
appropriate surrogate $\tilde{C}_t$ of $C_t$, whose minimizer coincides with the SGD 
update for $\bL[t]$ in~\eqref{eq:stochastic_gradient}. The surrogate $\tilde{C}_t$ 
then plays the same role as $\hat{C}_t$, associated with the second-order algorithm 
towards the convergence analysis. Recall that 
$\bq[t]=\arg\min_{\bq \in \mathbb{R}^{\rho}} g_t(\bL[t-1],\bq)$. In this direction, 
in the average cost $\hat{C}_t(\bL)=\frac{1}{t} \sum_{\tau=1}^t f_t(\bL,\bq[t])$ 
[cf. (P3) for $\theta=1$], with $f_t(\bL,\bq[t])=g_{t}(\bL,\bq[t])+\frac{\lambda_t}{2t} \|\bL\|_F^2$ 
one can further approximate $f_t$ using the second-order Taylor expansion at the previous 
subspace update $\bL[t-1]$. This yields
\begin{align}
\tilde{f}_t(\bL,\bq[t])=f_t(\bL[t-1],\bq[t]) + \tr\big\{\nabla_{\bL} 
f_t(\bL[t-1],\bq[t]) (\bL-\bL[t-1])'\big\} + \frac{\alpha_t}{2} \|\bL-\bL[t-1]\|_F^2 
\label{eq:tilde_g_t}
\end{align}
where $\alpha_t \geq \|\nabla^2 f_t(\bL[t-1],\bq[t])\| = 
\|(\bq[t]\bq'[t]) \otimes \bOmega_t + \lambda_t/2t \bI_{P\rho}\|$. 

It is useful to recognize that the surrogate $\tilde{f}_t$ is a tight approximation of 
$f_t$ in the sense that: (i) it globally majorizes the original cost function $f_t$, 
i.e., $\tilde{f}_t(\bL,\bq[t]) \geq f_t(\bL,\bq[t]),~\forall \:\bL \in \mathbb{R}^{P \times \rho}$; 
(ii) it is locally tight, namely $\tilde{f}_t(\bL[t-1],\bq[t])=f_t(\bL[t-1],\bq[t])$; and, (iii) 
its gradient is locally tight, namely $\nabla_{\bL} \tilde{f}_t(\bL[t-1],\bq[t])=\nabla_{\bL} 
f_t(\bL[t-1],\bq[t])$. Consider now the average approximate cost 
\begin{align}
\tilde{C}_t(\bL)=\frac{1}{t} \sum_{\tau=1}^t \tilde{f}_{\tau}(\bL,\bq[\tau])  \label{eq:tilde_C_t}
\end{align}
where due to (i) it follows that $\tilde{C}_t(\bL) \geq \hat{C}_t(\bL) \geq C_t(\bL)$ 
holds for all $\bL \in \mathbb{R}^{P \times \rho}$. The subspace update $\bL[t]$ is 
then obtained as $\bL[t]:=\arg\min_{\bL \in \mathbb{R}^{P \times \rho}} \tilde{C}_t(\bL)$, 
which amounts to nulling the gradient [cf. \eqref{eq:tilde_g_t} and \eqref{eq:tilde_C_t}]
\begin{align}
\nabla \tilde{C}_t(\bL[t]) = \frac{1}{t} \sum_{\tau=1}^t \Big\{\nabla_{\bL} 
f_{\tau} (\bL[\tau-1],\bq[\tau]) + \alpha_{\tau} (\bL - \bL[\tau-1]) \Big\}. \nonumber
\end{align}
After defining $\bar{\alpha}_t:=\sum_{\tau=1}^t \alpha_{\tau}$, the first-order optimality 
condition leads to the recursion
\begin{align}
\bL[t] &= \frac{1}{\bar{\alpha}_t} \sum_{\tau=1}^t  \alpha_{\tau} \Big(\bL[\tau-1] - 
\alpha_{\tau}^{-1} \nabla_{\bL} f_{\tau}(\bL[\tau-1],\bq[\tau]) \Big) \nonumber\\
&=\frac{1}{\bar{\alpha}_t} \underbrace{\sum_{\tau=1}^{t-1}  \alpha_{\tau} \Big(\bL[\tau-1] - 
\alpha_{\tau}^{-1} \nabla_{\bL} f_{\tau}(\bL[\tau-1],\bq[\tau]) \Big)}_{\normalfont{:=\bar{\alpha}_{t-1} 
\bL[t-1]}}  + \frac{\alpha_t}{\bar{\alpha}_t} \Big(\bL[t-1]-\alpha_t^{-1} 
\nabla_{\bL} f_t(\bL[t-1],\bq[t])\Big)  \nonumber\\
&= \bL[t-1] - \frac{1}{\bar{\alpha}_t} \nabla_{\bL} f_t(\bL[t-1],\bq[t]). \label{eq:first_order_recursion} 
\end{align}
Upon choosing the step size sequence $(\mu[t])^{-1}:=\bar{\alpha}_t^{-1}$, 
the recursion in~\eqref{eq:stochastic_gradient} readily follows. 

Now it only remains to verify that the main steps of the proof outlined under (S1) and (S2) in Section \ref{subsec:cnvg_secondorder_alg}, carry over for the average approximate cost~$\tilde{C}_t$. 
Under (A1)--(A3) and thanks to the approximation tightness of $\tilde{f}_t$ 
as reflected through (i)-(iii), one can follow the same arguments in the proof of
Proposition \ref{prop:prop_2} (see also~\cite[Lemma 3]{jstsp_anomalography_2012}) to show
that $\{\tilde{C}_t(\bL[t])\}$ is a quasi-martingale sequence, and $\lim_{t \rightarrow \infty} (\tilde{C}_t(\bL[t])-C_t(\bL[t]))=0$. Moreover, assuming the sequence $\{\alpha_t\}$ is 
bounded and under the compactness assumption (A3), the quadratic function $\tilde{f}_t$ 
fulfills the required regularity conditions (\cite[Lemma 1]{jstsp_anomalography_2012} so that 
(S2) holds true. All in all, the SGD algorithm is convergent as formalized in the following claim.

\begin{proposition}\label{prop:prop_4}
Under (A1)--(A3) and for $\lambda_t=\lambda~\forall t$, if $\mu[t]:=\sum_{\tau=1}^t \alpha_{\tau} \geq ct$ 
for some constant $c>0$ and $c' \geq \alpha_t \geq \|(\bq[t]\bq'[t]) \otimes \bOmega_t + 
\lambda/2t \bI_{P\rho}\|,~\forall t$ hold, the subspace iterates \eqref{eq:stochastic_gradient} 
satisfy $\lim_{t \rightarrow \infty} \nabla C_t(\bL[t]) = \mathbf{0}_{P\times\rho}$ a.s., i.e.,~$\{\bL[t]\}_{t=1}^\infty$ asymptotically coincides with the 
stationary points of the batch problem (P2).
\end{proposition}

\begin{remark}[Convergence of accelerated SGD]\label{rem:remark_3}
Paralleling the steps of the convergence proof for the SGD algorithm outline before, 
one may expect similar claims can be established for the \emph{accelerated} variant tabulated
under Algorithm \ref{tab:table_2}. However, it is so far not clear 
how to construct an appropriate surrogate $\tilde{C}_t$ based on the available subspace updates $\{\bL[t]\}$, 
whose minimizer coincides with the extrapolated estimates $\tilde{\bL}[t]$. %Therefore, the proof techniques adopted 
%based on the theory of martingale sequences is not easily applicable, and hence it is left as a 
%future research. 
Recently, a variation of the accelerated SGD algorithm was put forth in~\cite{kostas_icassp2014_cnvg},
which could be applicable to the subspace tracking problem studied in this paper. 
Adopting a different proof technique, the algorithm of~\cite{kostas_icassp2014_cnvg} is
shown convergent, and this methodology could be instrumental in formalizing the convergence
of Algorithm \ref{tab:table_2} as well.
\end{remark}

\subsection{Optimality}
\label{subsec:optimality}
Beyond convergence to stationary points of (P2), one may ponder whether the online estimator 
offers performance guarantees of the batch nuclear-norm 
regularized estimator (P1), for which stable/exact recovery results are well documented 
e.g., in~\cite{CR08,candes_moisy_mc}. Specifically, given the learned subspace $\bar{\bL}[t]$ 
and the corresponding $\bar{\bQ}[t]$ [obtained via~\eqref{eq:rec_est_q}] over a time window 
of size $t$, is $\{\hat{\bX}[t]:=\bar{\bL}[t]\bar{\bQ}'[t]\}$ an optimal solution
of (P1) as $t \rightarrow \infty$? This in turn requires asymptotic analysis
of the optimality conditions for (P1) and (P2), and a positive answer is established in the next proposition 
whose proof is deferred to the Appendix. Additionally, numerical tests in Section~\ref{sec:sims} 
indicate that Algorithm \ref{tab:table_1} attains the performance of (P1) after a modest
number of iterations.

\begin{proposition}\label{prop:prop_3}
Consider the subspace iterates $\{\bL[t]\}$ generated by either Algorithm~\ref{tab:table_1} (with $\theta=1$),
or Algorithm~\ref{tab:table_2}. If there exists a subsequence $\{\bL[t_k],\bQ[t_k]\}$ for which (c1) 
$\lim_{k \rightarrow \infty}$ $\nabla C_{t_k}(\bL[t_k]) $ $= \mathbf{0}_{P\times\rho}$ a.s., and 
(c2) $\frac{1}{\sqrt{t_k}}\sigma_{\max} [\cP_{\Omega_{t_k}}(\bY_{t_k}-\bL[t_k]\bQ'[t_k])]$ 
$\leq \frac{\lambda_{t_k}}{\sqrt{t_k}}$ hold, then the sequence $\{\bX[k]=\bL[t_k]\bQ'[t_k]\}$ 
satisfies the optimality conditions for (P1) [normalized by $t_k$] as $k \rightarrow \infty$ a.s.
\end{proposition}

Regarding  condition (c1), even though it holds for a time invariant rank-controlling 
parameter $\lambda$ as per Proposition~\ref{prop:prop_2}, numerical tests indicate that it still 
holds true for the time-varying case (e.g., when $\lambda_t$
is chosen as suggested in Remark~\ref{rem:remark_2}). Under (A2) and (A3) one has 
$\sigma_{\max} [\cP_{\Omega_{t}}(\bY_{t}-\bL[t]\bQ'[t])]$ $\approx \mathcal{O}(\sqrt{t})$, 
which implies that the quantity on the left-hand side of (c2) cannot grow unbounded. Moreover, 
upon choosing $\lambda_t \approx \mathcal{O}(\sqrt{t})$ as per Remark~\ref{rem:remark_2} the term 
in the right-hand side of (c2) will not vanish, which suggests that the qualification condition 
can indeed be satisfied.

\section{Online Tensor Decomposition and Imputation}
\label{sec:tensor_completion}

As modern and massive datasets become increasingly complex and
heterogeneous, in many situations one encounters data structures indexed
by three or more variables giving rise to a tensor, instead of just 
two variables as in the matrix settings studied so far. 
%As with matrices, it is not uncommon 
%that one of these variables indexes time~\cite{nion_online_tensor}, and that sizeable portions of the data are 
%missing~\cite{juan_tensor_tsp_2013,kolda_completion,GRY11}. 
A few examples of time-indexed, incomplete tensor
data include~\cite{kolda_completion}: (i) dynamic social networks represented through a temporal sequence
of network adjacency matrices, meaning a data cube with entries indicating whether e.g., two agents coauthor a paper or 
exchange emails during time interval $t$, while it may be the case that not all pairwise interactions can be sampled;  
(ii) Electroencephalogram (EEG) data, where each signal from an electrode can be represented as a
time-frequency matrix; thus, data from multiple channels is three-dimensional (temporal, spectral, 
and spatial) and may be incomplete if electrodes become loose or disconnected for a period of time; 
and (iii) multidimensional nuclear magnetic resonance (NMR) analysis,
where missing data are encountered when sparse sampling is used in order to reduce the experimental time.

Many applications in the aforementioned domains aim at capturing the 
underlying latent structure of the data, which calls for high-order factorizations 
even in the presence of missing data~\cite{juan_tensor_tsp_2013,kolda_completion}. 
%While it is in principle possible to unfold the 
%given tensor data into a matrix and resort to the algorithms in 
%Sections \ref{sec:onlineMC} and \ref{sec:low_compx_sub_update}, 
%tensor models preserve the multi-way nature of the data and extract the 
%underlying factors in each mode (dimension) of a higher-order array.  
Accordingly, the desiderata for analyzing streaming
and incomplete multi-way data are low-complexity, real-time algorithms
capable of unraveling latent structures through parsimonious (e.g., low-rank) decompositions,
such as the PARAFAC model described next. %Accurately approximating 
%a given incomplete tensor allows one
%to impute those missing entries as a byproduct, by simply reconstructing the data cube by means
%of the estimated model factors. 
In the sequel, the discussion will be focused on three-way
tensors for simplicity in exposition, but extensions to higher-way arrays are possible.

\subsection{Low-rank tensors and the PARAFAC decomposition}
\label{ssec:parafac}

For three vectors $\ba\in\mathbb{R}^{M\times 1}$, $\bb\in\mathbb{R}^{N\times 1}$, 
and $\bc\in\mathbb{R}^{T\times 1}$, the outer product $\ba \circ \bb \circ \bc $ is an 
$M\times N\times T$ rank-one three-way array with $(m,n,t)$-th entry given by 
$\ba(m) \bb(n) \bc(t)$. Note that this comprises a generalization of the two vector (matrix) 
case, where $\ba \circ \bb =\ba \bb'$ is a rank-one matrix. The rank of a tensor 
$\underline{\bX}$ is defined as the minimum number of outer products required to 
synthesize $\underline{\bX}$.

\begin{figure}[t]
\centering
\includegraphics[scale=0.6]{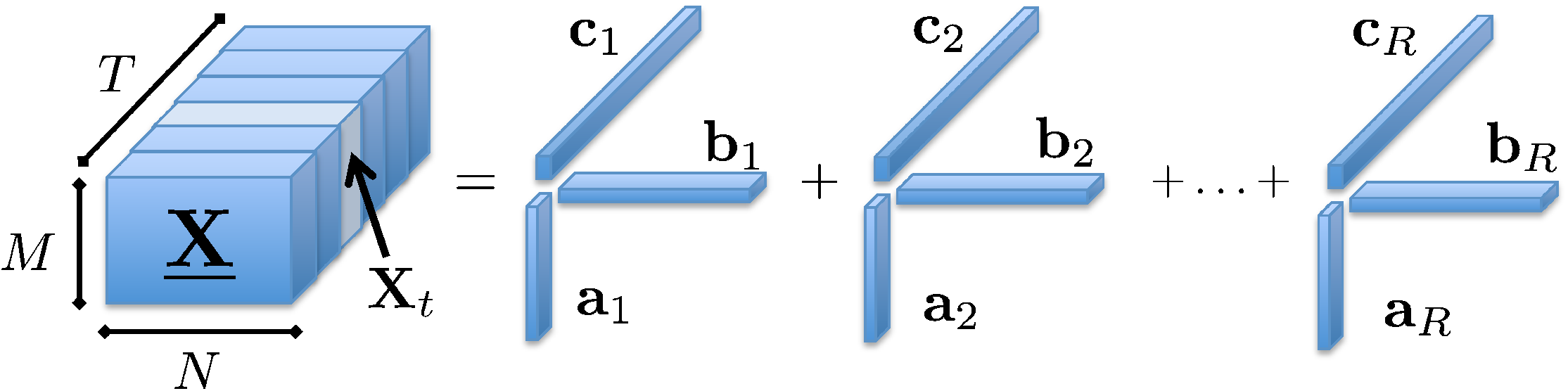}
\caption{A rank-$R$ PARAFAC decomposition of the three-way tensor $\underline{\mathbf{X}}$.}
\label{fig:Fig_1}
\end{figure}

The PARAFAC model is arguably the most basic tensor model because of its direct relationship to tensor
rank. Based on the previous discussion it is natural to form a \emph{low-rank approximation}
of tensor $\underline{\bX} \in \mathbb{R}^{M \times N \times T}$ as
\begin{align}
\underline{\bX} \approx \sum_{r=1}^R \ba_r \circ \bb_r \circ \bc_r . \label{eq:parafac}
\end{align}
When the decomposition is exact, \eqref{eq:parafac} is the PARAFAC decomposition of
$\underline{\bX}$; see also Fig. \ref{fig:Fig_1}. Accordingly, the minimum value $R$ for which the exact
decomposition is possible is (by definition) the rank of $\underline{\bX}$. 
PARAFAC is the model of choice when one is 
primarily interested in revealing latent structure. Considering the analysis of a
dynamic social network for instance, each of the rank-one factors in Fig. \ref{fig:Fig_1}
could correspond to communities that e.g., persist or form and dissolve periodically
across time.
Different from the matrix case, there is no straightforward algorithm
to determine the rank of a given tensor, a problem that has been shown to be NP-hard.
For a survey of algorithmic approaches to obtain approximate PARAFAC
decompositions, the reader is referred to~\cite{kolda_tutorial}. 

With reference to \eqref{eq:parafac}, introduce the factor matrix 
$\bA:=[\ba_1,\ldots, \ba_R ] \in \mathbb{R}^{M \times R}$, and likewise for 
$\bB \in \mathbb{R}^{N \times R}$ and $\bC \in \mathbb{R}^{T \times R}$.
Let  $\bX_t,\ t=1,\ldots,T$ denote the $t$-th slice of $\underline{\bX}$ along its 
third (tube) dimension, such that $\bX_t(m,n)=\underline{\bX}(m,n,t)$; see also Fig. \ref{fig:Fig_1}. 
The following compact matrix form of the PARAFAC
decomposition in terms of slice factorizations will be used in the sequel
\begin{align}
\bX_t = \bA {\rm diag}(\bxi_t) \bB' = \sum_{r=1}^R \bxi_t(r) \ba_r \bb_r',\quad t=1,2,\ldots,T \label{eq:slice_t}
\end{align}
where $\bxi_t'$ denotes the $t$-th row of $\bC$ (recall that $\bc_r$ instead denotes the 
$r$-th column of $\bC$). It is apparent that each slice $\bX_t$ can 
be represented as a linear combination of $R$ rank-one matrices $\{\ba_r\bb_r'\}_{r=1}^R$, 
which constitute the bases for the tensor fiber subspace. 
The PARAFAC decomposition is symmetric [cf. \eqref{eq:parafac}],
and one can likewise write $\bX_m=\bB {\rm diag}(\balpha_m) \bC'$,
or, $\bX_n=\bC {\rm diag}(\bbeta_n) \bA'$ in terms of slices
along the first (row), or, second (column) dimensions -- once more, $\balpha_m'$ stands for
the $m$-th row of $\bA$, and likewise for $\bbeta_n'$.
Given $\underline{\mathbf{X}}$, under some
technical conditions then $\{\bA,\bB,\bC\}$ are unique up to
a common column permutation and scaling (meaning PARAFAC is identifiable); 
see e.g.~\cite{tBS02,jk77laa}

%\begin{remark}[Low-rank decompositions for imputation]\label{rem:imputation_feasibility}
%\normalfont 

Building on the intuition for the matrix case, feasibility of the 
imputation task relies fundamentally on assuming a
low-dimensional PARAFAC model for the data, to couple the available and missing entries as well as
reduce the effective degrees of freedom in the problem. 
Under the low-rank assumption for instance, a rough idea on the fraction $p_m$ of missing data that 
can be afforded is obtained by comparing the number of unknowns $R (M+N+T)$ in \eqref{eq:parafac} 
with the number of available data samples  $(1-p_m)MNT$. Ensuring that $(1-p_m)MNT\geq R (M+N+T)$, 
roughly implies that the tensor can be potentially recovered even if a fraction 
$p_m\leq 1 - R (M+N+T)/(MNT)$ of entries is missing.
Different low-dimensional tensor models would lead to alternative imputation methods, 
such as the unfolded tensor regularization in \cite{GRY11,tensor_completion_visualdata_liu13} for \emph{batch} tensor completion. 
The algorithm in the following section offers (for the first time) an approach 
for decomposing and imputing low-rank \emph{streaming} tensors.

\subsection{Algorithm for streaming tensor data}
\label{ssec:tensor_algorithm}

\begin{figure}[t]
\centering
\includegraphics[scale=0.6]{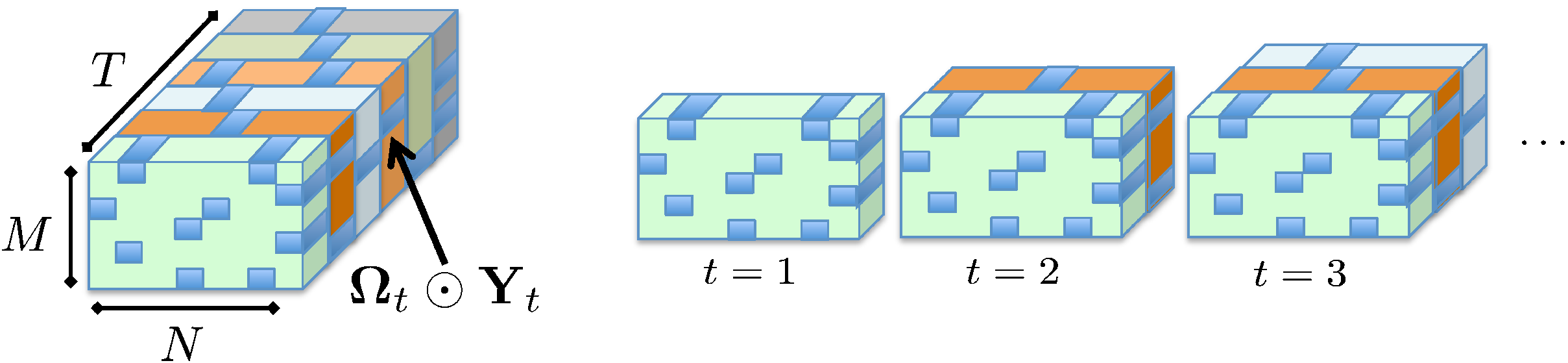}
\caption{Tensor data with missing entries. (Left) Batch data, and slice $\bm{\Omega}_t\odot\bY_t$ along
the time (tube) dimension. (Right) Streaming data, where slices $\bm{\Omega}_t\odot\bY_t$ become available
for $t=1,2,\ldots$.}
 \label{fig:Fig_2}
\end{figure}

Let $\underline{\bY}\in\mathbb{R}^{M\times N\times T}$ be a three-way tensor, and likewise let $\underline{\bOmega}$ 
denote a $M\times N \times T$ binary $\{0,1\}$-tensor with $(m,n,t)$-th entry equal to $1$ if 
$\underline{\bY}(m,n,t)$ is observed, and $0$ otherwise. One can thus represent the
incomplete data tensor compactly as $\mathcal{P}_{\underline{\bOmega}}(\underline{\bY})=
\underline{\bOmega}\odot \underline{\bY}$; see also Fig. \ref{fig:Fig_2} (left). Generalizing the nuclear-norm 
regularization technique in (P1) from low-rank matrix to tensor completion is not
straightforward if one also desires to unveil the latent structure in the data. The notion 
of singular values of a tensor (given by the Tucker3 decomposition) are not related
to the  rank~\cite{kolda_tutorial}. Interestingly, it was argued in~\cite{juan_tensor_tsp_2013} 
that the Frobenius-norm regularization outlined in Section \ref{subsec:separable} 
offers a viable option for \emph{batch} low-rank tensor completion under the PARAFAC model, by solving 
[cf. (P2) and \eqref{eq:slice_t}]
\begin{align}
{\rm (P4)}~~~~~\min_{\{\bX,\bA \in \mathbbm{R}^{M \times \hat{R}},\bB\in \mathbbm{R}^{N \times \hat{R}},\bC\in \mathbbm{R}^{T \times \hat{R}}\}} &\frac{1}{2}  \|\underline{\bOmega} \odot 
(\underline{\bY}-\underline{\bX})\|_F^2 + \frac{\lambda}{2} (\|\bA\|_F^2+\|\bB\|_F^2 +  
\|\bC\|_F^2)  \nonumber\\
& {\rm s.~to} ~~\bX_t =\bA {\rm diag}(\bxi_t) \bB',\quad t=1,2\ldots,T. \nonumber%\label{eq:batch_formulation}
\end{align}
The regularizer in (P4) provably
encourages low-rank tensor decompositions, in fact with controllable rank by tuning the 
parameter $\lambda$~\cite{juan_tensor_tsp_2013}. Note that similar to the matrix case there is no need for the true rank $R$ in (P4). In fact, any upperbound $\hat{R} \geq R$ can be used for the column size of the sought matrix variables $\bA,\bB,\bC$ as long as $\lambda$ is tuned appropriately.

Consider now a real-time setting where the incomplete tensor slices 
$\bm\Omega_t\odot\bY_t$ are acquired sequentially over time
$t=1,2,\ldots$ [i.e., streaming data as depicted in Fig. \ref{fig:Fig_2} (right)]. Leveraging 
the batch formulation (P4) one can naturally broaden the 
subspace tracking framework in Section \ref{sec:onlineMC}, to devise adaptive 
algorithms capable of factorizing tensors `on the fly'. To this end, one can
estimate the PARAFAC model factors $\{\bA[t],\bB[t],\bC[t]\}$ as the minimizers of
the following EWLS cost [cf. (P3)]
\begin{align}
{\rm (P5)}~~\min_{\{\bA,\bB,\bC\}} \frac{1}{2} \sum_{\tau=1}^t \theta^{t-\tau} 
\Big[\|\bOmega_{\tau} \odot (\bY_{\tau}-\bA {\rm diag}(\bxi_{\tau})\bB')\|_F^2 + 
\bar{\lambda}_t (\|\bA\|_F^2+\|\bB\|_F^2) + \lambda_t \|\bxi_{\tau}\|^2 \Big]. \nonumber%\label{eq:online_formulation}
\end{align}
Once more, the normalization  $\bar{\lambda}:=\lambda_t/\sum_{\tau=1}^t \theta^{t-\tau}$ ensures that
for the infinite memory setting ($\theta=1$) and $t=T$, (P5) coincides with the batch estimator (P4).

Paralleling the algorithmic construction steps adopted for the matrix case, upon defining 
the counterpart of $g_t(\bL,\bq)$
corresponding to (P5) as
\begin{align}
\bar{g}_t(\bA,\bB,\bxi):=\frac{1}{2} \|\bOmega_t \odot (\bY_t-\bA {\rm diag}(\bxi)\bB')\|_F^2 + 
\frac{\lambda_t}{2} \|\bxi\|^2 \label{eq:fun_g_t}
\end{align}
the minimizer $\bxi_t=\arg\min_\bxi\bar{g}_t(\bA,\bB,\bxi) $ is readily obtained in closed form, namely
\begin{align}
\bxi_t= \left[ \lambda \bI_R + \sum_{(m,n) \in \Omega_t} (\balpha_m\odot \bbeta_n)(\balpha_m \odot 
\bbeta_n)' \right]^{-1} \sum_{(m,n) \in \Omega_t} \underline{\bY}_t(m,n)(\balpha_m\odot\bbeta_n ).
\label{eq:xi_coeffs}
\end{align}
Accordingly, the factor matrices $\{\bA,\bB \}$ that can be interpreted as bases for the fiber 
subspace are the minimizers
of the cost function
\begin{align}
\bar{C}_t(\bA,\bB ) := \sum_{\tau=1}^t \theta^{t-\tau} \bar{g}_{\tau}(\bA,\bB,\bxi_{\tau}) + 
\frac{\bar{\lambda}_t}{2} (\|\bA\|_F^2+\|\bB\|_F^2). \label{eq:online_form_g_t}
\end{align}
Note that $\bxi_t:=\bxi_t(\bA,\bB)$ as per \eqref{eq:xi_coeffs}, so minimizing $\bar{C}_t(\bA,\bB )$  becomes 
increasingly complex computationally as $t$ grows.
\begin{remark}[Challenges facing a second-order algorithm]\label{rem:challenges_second_order}
\normalfont As discussed in Section \ref{sec:onlineMC}, one can approximate $\bar{g}_{t}(\bA,\bB,\bxi_t)$ 
with the upper bound $\bar{g}_{t}(\bA,\bB,\bxi_t(\bA[t-1],\bB[t-1]))$ to develop a second-order algorithm that 
circumvents the aforementioned increasing complexity roadblock. Unlike the 
matrix case however, \eqref{eq:online_form_g_t} is a nonconvex problem due to the bilinear nature of the PARAFAC
decomposition (when, say, $\bC$ is fixed); thus, finding its global optimum efficiently is challenging. 
One could instead think of carrying out alternating minimizations with respect to each of the tree factors 
per time instant $t$, namely updating: (i) $\bxi[t]$ first, given $\{\bA[t-1],\bB[t-1]\}$; (ii) then $\bB[t]$ 
given $\bA[t-1]$ and $\{\bxi[\tau]\}_{\tau=1}^t$; and (iii) finally $\bA[t]$ with fixed $\bB[t]$ and 
$\{\bxi[\tau]\}_{\tau=1}^t$. While each of these subtasks boils down to a convex optimization problem, 
the overall procedure does not necessarily lead to an efficient algorithm since one can show
that updating $\bA[t]$ and $\bB[t]$ recursively is impossible.
\end{remark}

Acknowledging the aforementioned challenges and the desire of computationally-efficient
updates compatible with Big Data requirements, it is prudent to seek instead a (first-order) SGD
alternative. Mimicking the steps in Section \ref{subsec:low_compx_sub_update}, let $\bar{f}_t(\bA,\bB):=\bar{g}_t(\bA,\bB,\bxi[t])+
\frac{\lambda_t}{2t} (\|\bA\|_F^2+\|\bB\|_F^2)$ denote the $t$-th summand in \eqref{eq:online_form_g_t}, for $t=1,2,\ldots$
and $\theta=1$. The factor matrices $\mathcal{L}[t]:=\{\bA[t],\bB[t]\}$ are obtained via the SGD iteration
\begin{equation}\label{eq:stochastic_gradient_tensor}
\mathcal{L}[t]=\arg\min_{\mathcal{L}}\bar{Q}_{\bar{\mu}[t],t}(\mathcal{L},\mathcal{L}[t-1])=\mathcal{L}[t-1]-(\bar{\mu}[t])^{-1}\nabla \bar{f}_t(\mathcal{L}[t-1])
\end{equation}
with the stepsize $(\bar{\mu}[t])^{-1}$, and $\bar{Q}_{\mu,t}(\mathcal{L}_1,\mathcal{L}_2):=\bar{f}_t(\mathcal{L}_2) + \langle\mathcal{L}_1-\mathcal{L}_2,\nabla
\bar{f}_t(\mathcal{L}_2)\rangle +\frac{\mu}{2}\|\mathcal{L}_1-\mathcal{L}_2\|^2$. It is instructive to recognize 
that the quadratic surrogate $\bar{Q}_{\bar{\mu}[t],t}$ has the following properties: (i) it majorizes $\bar{f}_t(\cL)$, namely 
$\bar{f}_t(\mathcal{L})\leq \bar{Q}_{\bar{\mu},t}(\mathcal{L},\mathcal{L}[t-1]),~\forall \cL$; while it is locally tight 
meaning that (ii)~$\bar{f}_t(\mathcal{L}[t-1])= \bar{Q}_{\bar{\mu}[t],t}(\mathcal{L}[t-1],\mathcal{L}[t-1])$, 
and (iii)~$\nabla \bar{f}_t(\mathcal{L}[t-1]) = \nabla \bar{Q}_{\bar{\mu}[t],t}(\mathcal{L}[t-1],\mathcal{L}[t-1])$. Accordingly, the minimizer of $\bar{Q}_{\bar{\mu}[t],t}(\mathcal{L},\mathcal{L}[t-1])$ amounts to  
a correction along the negative gradient $\nabla \bar{f}_t(\mathcal{L}[t-1])$, with  stepsize $(\bar{\mu}[t])^{-1}$ 
[cf. \eqref{eq:stochastic_gradient_tensor}]. 

\begin{algorithm}[t]
\caption{: Online SGD algorithm for tensor decomposition and imputation} \small{
\begin{algorithmic}
	\STATE \textbf{input} $\{\bY_t,\bm{\Omega}_t\}_{t=1}^\infty,$, $\{\bar{\mu}[t]\}_{t=1}^{\infty}$, $\hat{R}$, and $\lambda_t$.
    \STATE \textbf{initialize} $\{\bA[0],\bB[0]\}$ at random, and $\bar{\mu}[0]>0$.
    \FOR {$t=0,1,2$,$\ldots$}
                \STATE $\bA'[t]:=[\balpha_1[t],\ldots,\balpha_M[t]]$ and $\bB'[t]:=[\bbeta_1[t],\ldots,\bbeta_N[t]]$
                \STATE $\bxi[t] = \left[ \lambda \bI_R + \sum_{(m,n) \in \Omega_t} (\balpha_m[t] \odot \bbeta_n[t])(\balpha_m[t] \odot \bbeta_n[t])' \right]^{-1} \sum_{(m,n) \in \Omega_t} \bY_t(m,n)( \balpha_m[t]\odot\bbeta_n[t] )$
                \STATE $\bA[t+1] = (1- \frac{\lambda_t}{t\bar{\mu}[t]})\bA[t] + \frac{1}{\bar{\mu}[t]} [\bOmega_t \odot (\bY_t-\bA[t] {\rm diag}(\bxi[t]) \bB'[t])] \bB[t] {\rm diag}(\bxi[t]) $
                \STATE $\bB[t+1] = (1- \frac{\lambda_t}{t\bar{\mu}[t]})\bB[t] + \frac{1}{\bar{\mu}[t]}[\bOmega_t \odot (\bY_t-\bA[t] {\rm diag}(\bxi[t]) \bB'[t])]' \bA[t] {\rm diag}(\bxi[t])$
    \ENDFOR
    \RETURN $\hat{\bX}[t]:=\bA[t] {\rm diag} (\bxi[t]) \bB'[t]$.
\end{algorithmic}}
\label{tab:table_3}
\end{algorithm}

Putting together \eqref{eq:xi_coeffs}  and \eqref{eq:stochastic_gradient_tensor}, while 
observing that the components of $\nabla
\bar{f}_t(\mathcal{L})$ are expressible as
\begin{align}
\nabla_{\bA} \bar{f}_t(\bA,\bB) &= -[\bOmega_t \odot (\bY_t-\bA {\rm diag}(\bxi[t]) \bB')] \bB {\rm diag}(\bxi[t]) + \frac{\lambda_t}{t}\bA \label{eq:grad_f_A_matrix}\\
\nabla_{\bB} \bar{f}_t(\bA,\bB) &= -[\bOmega_t \odot (\bY_t-\bA {\rm diag}(\bxi[t]) \bB')]' \bA {\rm diag}(\bxi[t]) + \frac{\lambda_t}{t}\bB \label{eq:grad_f_B_matrix} 
\end{align}
one arrives at the SGD iterations tabulated under Algorithm~\ref{tab:table_3}. Close examination of the recursions reveals that updating $\bA[t]$ and $\bB[t]$ demands $\mathcal{O}(|\Omega_t|\hat{R})$ 
operations, while updating $\bxi[t]$ incurs a cost of $\mathcal{O}(|\Omega_t|\hat{R}^2)$. The overall complexity per iteration 
is thus $\mathcal{O}(|\Omega_t|\hat{R}^2)$. 

\begin{remark}[Forming the tensor decomposition `on-the-fly']\label{rem:onthefly}
In a stationary setting the low-rank tensor decomposition can be accomplished 
after the tensor subspace matrices are learned; that is, when the sequences $\{\bA[t],\bB[t]\}$
converge to the limiting points, say $\{\bar{\bA},\bar{\bB}\}$. The remaining factor 
$\bar{\bC}:=[\bar{\bm{\gamma}}_1',\ldots,\bar{\bm{\gamma}}_T']'$ is then obtained by solving 
$\bar{\bm{\gamma}}_t=\arg\min_{\bm\gamma}\bar{g}_t(\bar{\bA},\bar{\bB},\bm{\gamma})$ for the corresponding
tensor slice $\calP_{\Omega_t}(\bX_t)$, which yields a simple closed-form solution as in \eqref{eq:xi_coeffs}. 
This requires revisiting the past tensor slices. The factors $\{\bar{\bA},\bar{\bB},\bar{\bC}\}$ 
then form a low-rank approximation of the entire tensor $\underline{\bX}\in\mathbb{R}^{M\times N\times T}$.
Note also that after the tensor subspace is learned say at time $t'\leq T$, e.g., from some
initial training data, the projection coefficients $\bar{\bm{\gamma}}_t$ can be calculated `on-the-fly' 
for $t\geq t'$; thus, Algorithm \ref{tab:table_3} offers a decomposition of then tensor containing slices $t'$
to $t>t'$ `on-the-fly'.
\end{remark}

Convergence of Algorithm~\ref{tab:table_3} is formalized in the next proposition, and can be established using similar arguments as in the matrix case detailed in Section~\ref{subsec:cnvg_stoc_gradient}. Furthermore, empirical observations in Section \ref{sec:sims} suggest that the convergence rate can be linear.

\begin{proposition}
Suppose slices $\{\bOmega_t \odot \bY_t\}_{t=1}^{\infty}$ and the corresponding sampling sets $\{\Omega_t\}_{t=1}^{\infty}$ 
are i.i.d., and $\theta=1$ while $\lambda_t=\lambda,~\forall t$. If (c1) $\{\cL[t]\}_{t=1}^{\infty}$ live in a compact set, and (c2) the step-size sequence $\{(\bar{\mu}[t])^{-1}\}$ satisfies $\bar{\mu}[t]:=\sum_{\tau=1}^t \tilde{\alpha}[\tau] \geq ct,~\forall t$ for some $c>0$, where (c3) $c' \geq \tilde{\alpha}[t] \geq \sigma_{\max}(\nabla^2 \bar{f}_t(\mathcal{L}[t-1])),~\forall t$ for some $c'>0$, then~$\lim_{t \rightarrow \infty} \nabla C_t(\cL[t])=\mathbf{0}$, a.s.; i.e., the tensor subspace iterates $\{\cL[t]\}$ asymptotically coincide with the stationary points of (P4).
\end{proposition}
%

% % % % % % % % % % % % % % % % % % % % % % % % % % % % % % % % % % % % % % % %
%                         Section V                                           %
% % % % % % % % % % % % % % % % % % % % % % % % % % % % % % % % % % % % % % % %

\section{Numerical Tests}
\label{sec:sims}
The convergence and effectiveness of the proposed algorithms is assessed in 
this section via computer simulations. Both synthetic and real data tests are 
carried out in the sequel.

\subsection{Synthetic matrix data tests} 

The signal $\bx_t=\bU\bw_t$ is generated from the low-dimensional subspace $\bU \in \mathbb{R}^{P \times r}$, 
with Gaussian i.i.d. entries $u_{p,i} \sim \cN(0,1/P)$, and projection coefficients 
$w_{i,t} \sim \cN(0,1)$. The additive noise $v_{i,t} \sim \cN(0,\sigma^2)$ is i.i.d., and to 
simulate the misses per time $t$, the sampling vector $\boldsymbol{\omega}_t \in \{0,1\}^{P}$ 
is formed, where each entry is a Bernoulli random variable, taking value one with probability 
(w.p.) $\pi$, and zero w.p. $1-\pi$, which implies that $(1-\pi)\times 100\%$ entries are
missing. The observations at time $t$ are generated as $\cP_{\omega_t}(\by_t) =
\boldsymbol{\omega}_t \odot (\bx_t+\bv_t)$. 

\begin{figure}[t]
\centering
\begin{tabular}{cc}
\hspace{-4mm}\epsfig{file=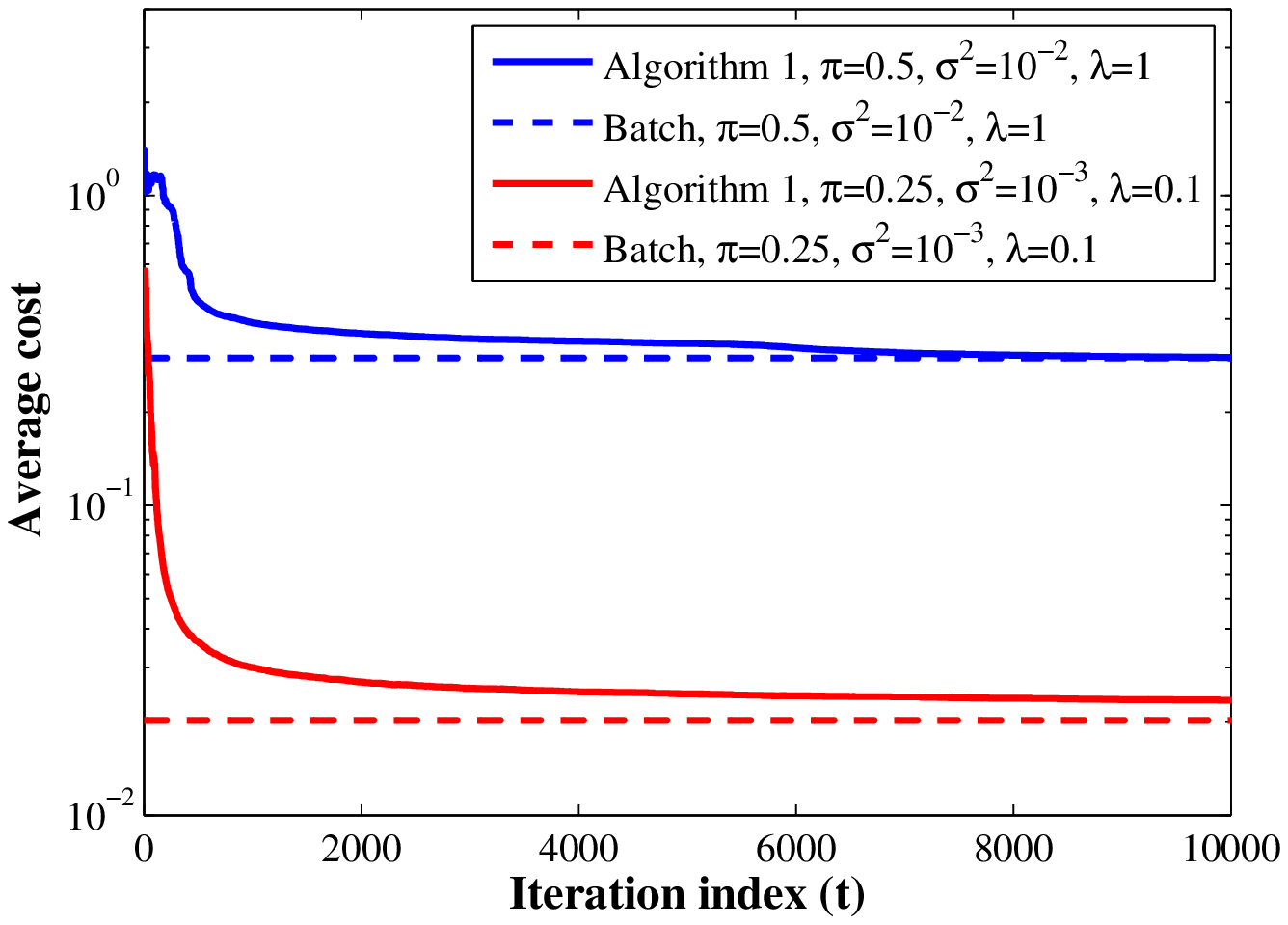,width=0.55
\linewidth, height=2.3 in } & \hspace{-8mm}
\epsfig{file=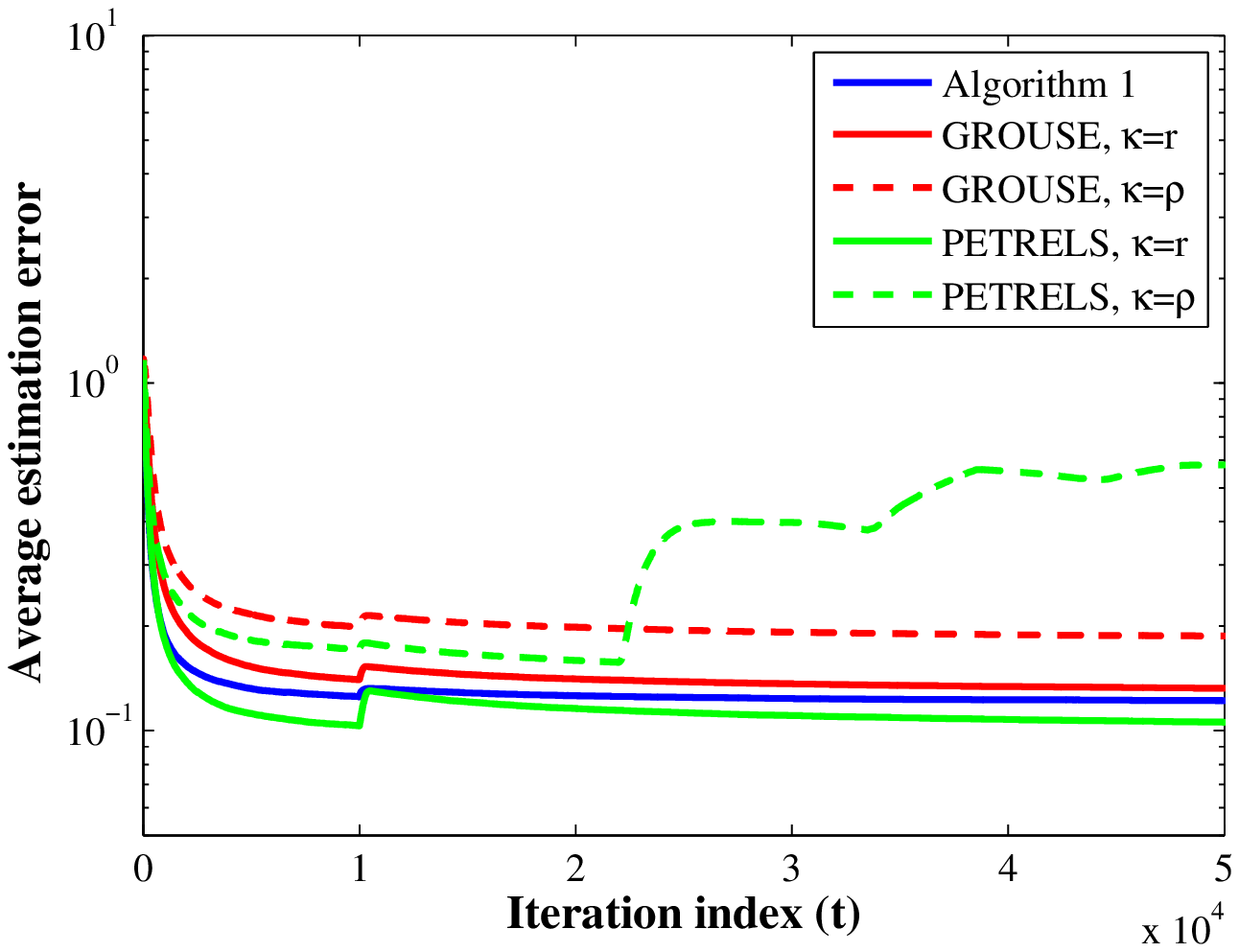,width=0.55
\linewidth, height=2.3 in } \\
(a) &
(b) \\
 \end{tabular}
 \vspace{-5mm}
 \caption{Performance of Algorithm~\ref{tab:table_1}. (a) Evolution of the average cost $C_t(\bL[t])$ versus the batch counterpart. (b) Relative estimation error for different schemes when $\pi=0.25$ and $\sigma^2=10^{-3}$.}
 \label{fig:fig_perf_synthdata}
\end{figure}

Throughout, fix $r=5$ and $\rho=10$, while different values of $\pi$ and $\sigma$ are examined. 
The time evolution of the average cost $C_t(\bL[t])$ in~\eqref{eq:cost_target} for 
various amounts of misses and noise strengths is depicted in 
Fig.~\ref{fig:fig_perf_synthdata}(a) [$\theta=1$]. For validation purposes, the optimal cost [normalized by the 
window size $t$] of the {\em batch} estimator (P1) is also shown. It is apparent that $C_t(\bL[t])$ 
converges to the optimal objective of the nuclear-norm regularized problem (P1), corroborating that 
Algorithm~\ref{tab:table_1} attains the performance of (P1) in the long run. This observation 
in addition to the low cost of Algorithm~\ref{tab:table_1}~[$\mathcal{O}(|\omega_t| \rho^2)$ per iteration] 
suggest it as a viable alternative for solving large-scale matrix completion problems.

Next, Algorithm~\ref{tab:table_1} is compared with other state-of-the-art subspace 
trackers, including PETRELS~\cite{petrels_chi12} and GROUSE~\cite{onlinetracking_bolzano10}, discussed in Section~\ref{sec:intro}. In essence, these algorithms need the dimension 
of the underlying subspace, say $\kappa$, to be known/estimated a priori. Fix $\lambda=0.1$, 
$\theta=0.99$, and introduce an abrupt subspace change at time $t=10^4$ to assess the tracking 
capability of the algorithms. The figure of merit depicted in Fig.~\ref{fig:fig_perf_synthdata}(b) 
is the running-average estimation error 
$e_x[t]:=\frac{1}{t} \sum_{i=1}^t \|\hat{\bx}_i - \bx_i\|_2 / \|\bx_i\|_2$.  
It is first observed that upon choosing identical 
subspace dimension $\kappa=\rho$ for all three schemes, Algorithm~\ref{tab:table_1} attains 
a better estimation accuracy, where a constant step size $(\mu[t])^{-1}=0.1$
was adopted for PETRELS and GROUSE. 
Albeit PETRELS performs well when the true rank is known, namely $\kappa=r$, 
if one overestimates the rank the algorithm exhibits erratic
behaviors for large fraction $75\%$ of missing observations. As expected, for the 
ideal choice of $\kappa=r$, all three schemes achieve nearly identical estimation 
accuracy. The smaller error exhibited by PETRELS relative to Algorithm~\ref{tab:table_1} 
may pertain to the suboptimum selection of $\lambda$. Nonetheless, for large amount 
of misses both GROUSE and PETRELS are numerically unstable as the LS problems to 
obtain the projection coefficients $\bq_t$ become ill-conditioned, whereas the 
ridge-regression type regularization terms in (P3) render 
Algorithm~\ref{tab:table_1} numerically stable. The price paid by Algorithm~\ref{tab:table_1} is however in terms of higher computational complexity per iteration, as seen in~Table~\ref{tab:table_4} which compares the complexity of various algorithms.

%Lastly, satisfaction of the 
%global optimality certificate for stationary points of (P2) is verified next. Fig.~\ref{fig:fig_qualification_vs_lambda}
%shows the qualification constraint violation $\sigma_{\max}[\cP_{\Omega_t}(\bY_t-\bar{\bL}_t\bar{\bQ}_t)]-\lambda$ 
%versus $\lambda$, for the noise level $\sigma^2=10^{-2}$, under $90\%$ misses and $r=3$. 
%For the given statistical data profile, when $\lambda$ exceeds a certain threshold,~the converged solution (which may be far from zero) is globally optimal. Note that 
%the corresponding threshold becomes smaller as $\rho$ increases, and the optimality conditions
%of (P1) show that the constraint violation is always nonnegative (and hence zero when stationary
%points are globally optimal). 
%This indicates that upon picking a large $\rho$, the converged solution of the 
%online Algorithm~\ref{tab:table_1} is likely to coincide with the global optimum of the 
%batch nuclear-norm regularized problem (P1), whose stable recovery performance is 
%guaranteed; see e.g.,~\cite{candes_moisy_mc}.

%%%%%%%%%%%%%%%%%%%%%%%%%%%%%%%%%%%%%%%%%%%%%%%%%%%%%%%%%%%%%%%%%%%%%%%%%%%%%%%
\begin{table*}[t]
\caption{Computational complexity per iteration}.
\vspace{-1cm}
\label{tab:table_4}
\begin{center}
\begin{tabular} {|c|c|c|c|c|c|c|c|}
\hline
 GROUSE $(\kappa=\rho)$ & PETRELS~($\kappa=\rho$) & Alg.~\ref{tab:table_1} ($\theta \in (0,1)$) & Alg.~\ref{tab:table_1} ($\theta=1$) & Alg.~\ref{tab:table_2} & Alg.~\ref{tab:table_2}~($k[t]=1$) \\
\hline
 $\mathcal{O}(P\rho+|\omega_t|\rho^2)$ & $\mathcal{O}(|\omega_t|\rho^2)$ & $\mathcal{O}(P\rho^3)$ & $\mathcal{O}(|\omega_t|\rho^2)$ & $\mathcal{O}(P\rho+|\omega_t| \rho^2)$ & $\mathcal{O}(|\omega_t|\rho^2)$\\
\hline
\end{tabular}
\end{center}
\end{table*}
%%%%%%%%%%%%%%%%%%%%%%%%%%%%%%%%%%%%%%%%%%%%%%%%%%%%%%%%%%%%%%%%%%%%%%%%%%%%%%%

%%%%%%%%%%%%%%%%%%%%%%%%%%%%%%%%%%%%%%%%%%%%%%%%%%%%%%%%%%%%%%%%%%%%%%%%%%%%%%%
%%%%%%%%%%%%%%%%%%%%%%%%%%%%%%%%%%%%%%%%%%%%%%%%%%%%%%%%%%%%%%%%%%%%%%%%%%%%%%%

%%%%%%%%%%%%%%%%%%%%%%%%%%%%%%%%%%%%%%%%%%%%%%%%%%%%%%%%%%%%%%%%%%%%%%%%%%%%%%%%
%\begin{figure}[t]
%\centering
%\begin{tabular}{cc}
%\hspace{-4mm}\epsfig{file=./figures/fig_qualification_vs_lambda.eps,width=0.6
%\linewidth, height=2.5 in }  \
% \end{tabular}
% \vspace{-5mm}
% \caption{Qualification constraint $\sigma_{\max}[\cP_{\Omega_t}(\bY_t-\bar{\bL}_t\bar{\bQ}'_t)] - \lambda_t$ under various values of $\rho$ when $P=T=100$, $r=3$, $\pi=0.1$, and $\sigma^2=10^{-2}$.}
% \label{fig:fig_qualification_vs_lambda}
%\end{figure}
%%%%%%%%%%%%%%%%%%%%%%%%%%%%%%%%%%%%%%%%%%%%%%%%%%%%%%%%%%%%%%%%%%%%%%%%%%%%%%%%

\subsection{Real matrix data tests} 
\label{subsec:real_matrix_test}

Accurate estimation of origin-to-destination (OD) flow traffic in the backbone of large-scale 
Internet Protocol (IP) networks is of paramount importance for proactive network security and 
management tasks~\cite{kolackzyk_book}. Several experimental studies have demonstrated that OD flow 
traffic exhibits a low-intrinsic dimensionality, mainly due to common temporal patterns across 
OD flows, and periodic trends across time~\cite{LPC04}. 
However, due to the massive number of OD pairs and the high 
volume of traffic, measuring the traffic of all possible OD flows is impossible for all
practical purposes~\cite{LPC04,kolackzyk_book}. 
Only the traffic level for a small fraction of OD flows can be measured via the NetFlow 
protocol~\cite{LPC04}. 

\begin{figure}[t]
\centering
\begin{tabular}{cc}
\hspace{-4mm}\epsfig{file=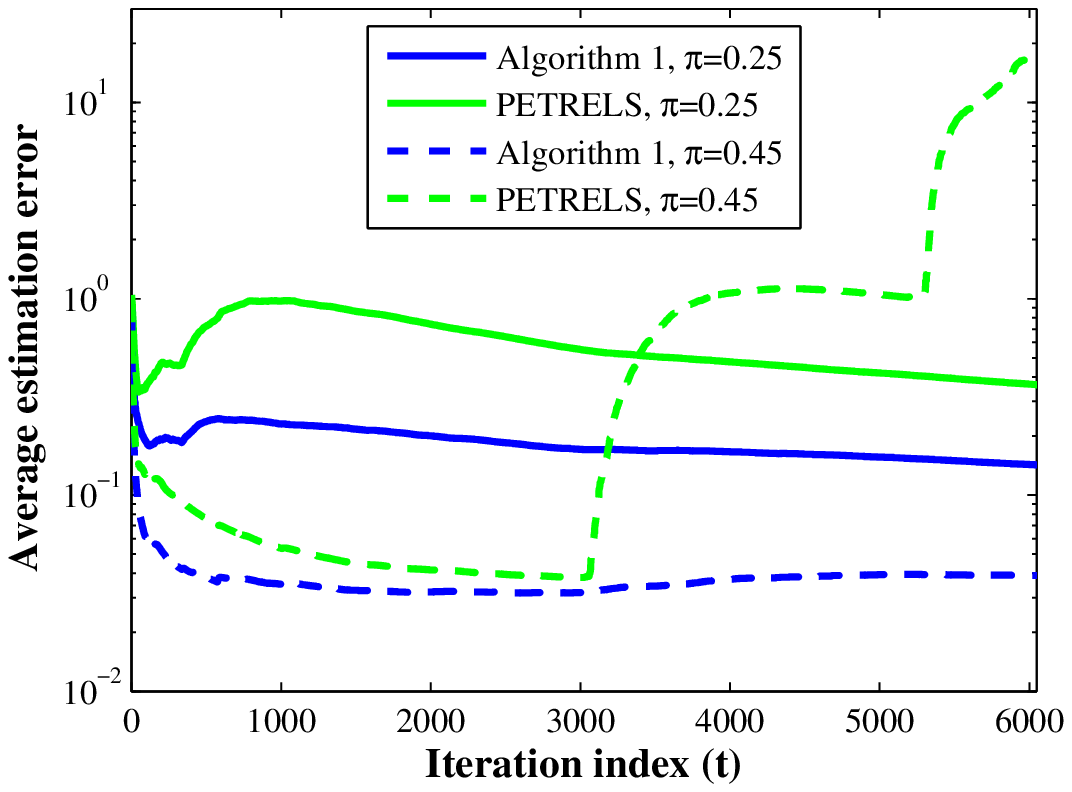,width=0.55
\linewidth, height=2.3 in } & \hspace{-8mm}
\epsfig{file=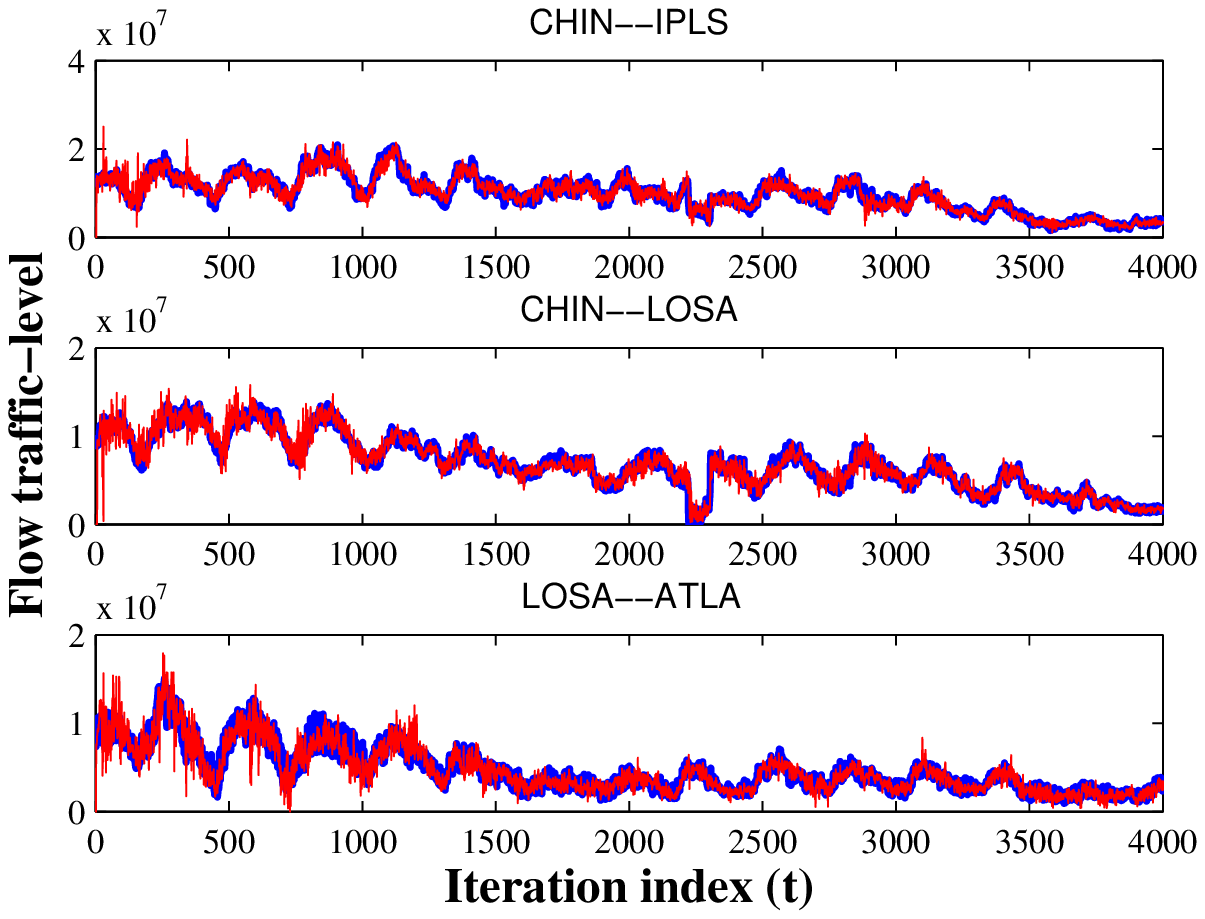,width=0.55
\linewidth, height=2.3 in } \\
(a) &
(b) \\
 \end{tabular}
 \vspace{-5mm}
 \caption{Traffic estimation performance for Internet-2 data when $\kappa=\rho=10$ and $\theta=0.95$. 
 (a) Average estimation error for various amounts of missing data. 
 (b) Algorithm \ref{tab:table_1}'s estimated (red) versus true (blue) OD flow traffic for $75\%$ misses ($\pi=0.25$).}
 \label{fig:fig_perf_realdata}
\end{figure}

Here, aggregate OD-flow traffic is collected from the operation of the Internet-2 
network~(Internet backbone across USA) during December 8-28, 2003 containing $121$ 
OD pairs~\cite{Internet2}. The measured OD flows contain spikes (anomalies), which are 
discarded to end up with a anomaly-free data stream $\{\by_t\} \in \mathbb{R}^{121}$. 
The detailed description of the considered dataset can be found in~\cite{jstsp_anomalography_2012}. 
A subset of entries of $\by_t$ are then picked randomly with probability $\pi$ to yield the 
input of Algorithm~\ref{tab:table_1}. The evolution of the running-average traffic estimation 
error ($e_x[t]$) is depicted in Fig.~\ref{fig:fig_perf_realdata}(a) for different schemes and
under various amounts of missing data. Evidently, Algorithm~\ref{tab:table_1} outperforms 
the competing alternatives when $\lambda_t$ is tuned adaptively as per Remark~\ref{rem:remark_2} 
for $\sigma^2=0.1$. When only $25\%$ of the total OD flows are sampled by Netflow, 
Fig.~\ref{fig:fig_perf_realdata}(b) depicts how Algorithm~\ref{tab:table_1} accurately 
tracks three representative OD flows. %Interestingly, GROUSE, tested under various step 
%sizes, performs very poorly in tracking the traffic subspace.

%%%%%%%%%%%%%%%%%%%%%%%%%%%%%%%%%%%%%%%%%%%%%%%%%%%%%%%%%%%%%%%%%%%%%%%%%%%%%%%
%%%%%%%%%%%%%%%%%%%%%%%%%%%%%%%%%%%%%%%%%%%%%%%%%%%%%%%%%%%%%%%%%%%%%%%%%%%%%%%

\subsection{Synthetic tensor data tests}

\begin{figure}[t]
\centering
\begin{tabular}{cc}
\hspace{-4mm}\epsfig{file=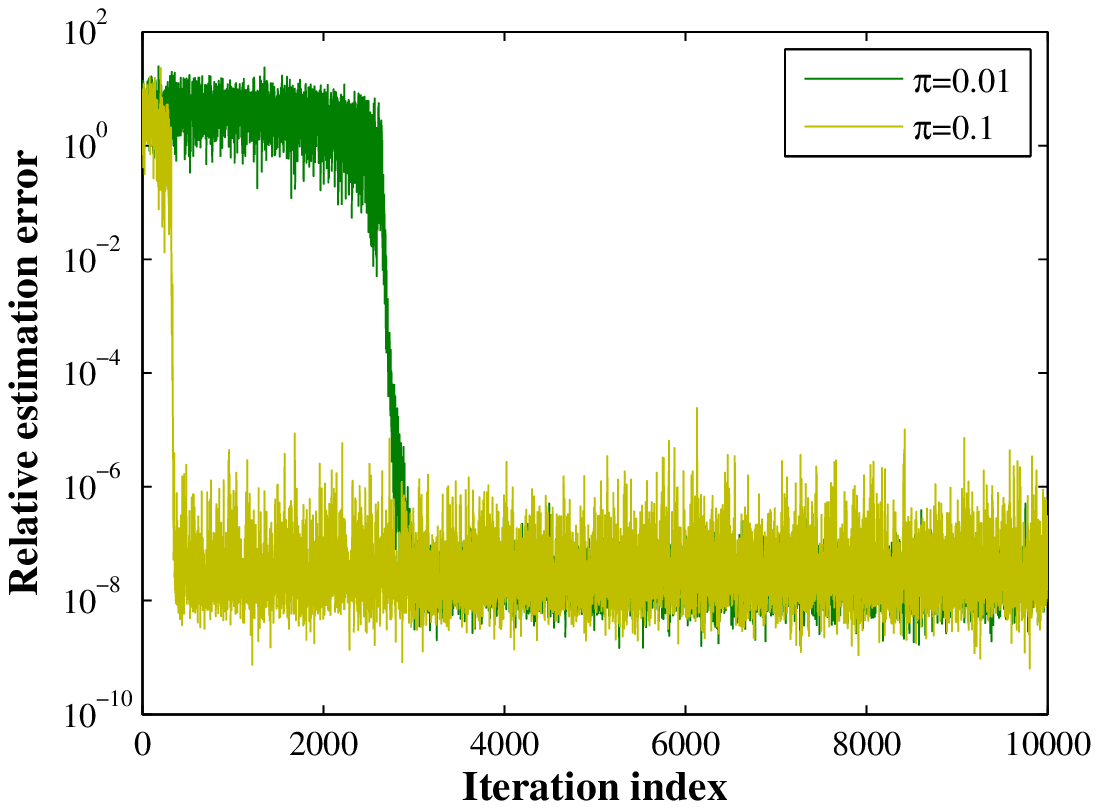,width=0.5
\linewidth, height=2.3 in } & \hspace{-8mm}
\epsfig{file=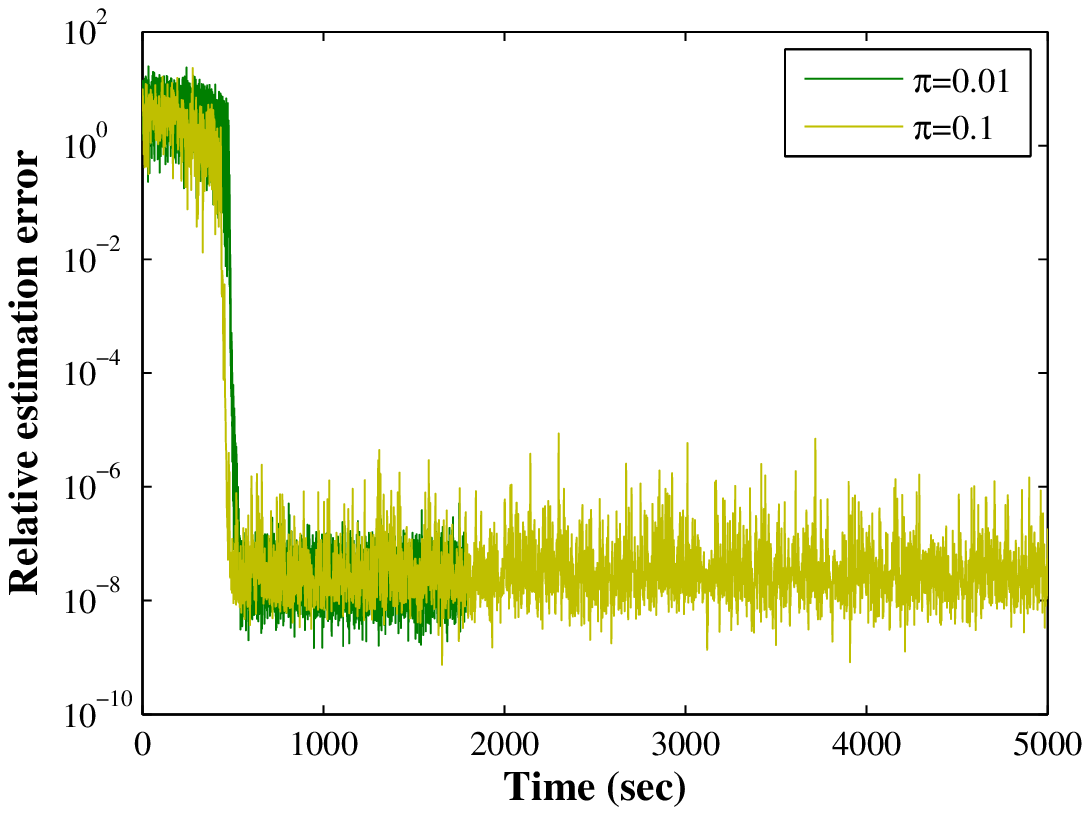,width=0.5
\linewidth, height=2.3 in } \\
(a) &
(b) \\
 \end{tabular}
 \vspace{-5mm}
 \caption{Relative estimation error versus (a) iterations and (b) run-time under various amounts of 
 misses for $M=N=10^3$, $\sigma=10^{-3}$ and $R=10$.}
 \label{fig:fig_perf_synthdata_error}
\end{figure}

To form the $t$-th `ground truth' tensor slice $\bX_t = \bA {\rm diag}(\bxi_t) \bB' \in 
\mathbb{R}^{M \times N}$, the factors $\bA$ and $\bB$ are generated independently with 
Gaussian i.i.d. columns $\ba_r \sim \calN(\mathbf{0},\bI_M)$ and $\bb_r \sim \calN(\mathbf{0},\bI_N)$; 
likewise, the coefficients $\bxi_t \sim \calN(0,\bI_{R})$. The sampling matrix $\bOmega_t$ also 
contains random Bernoulli entries taking value one w.p. $\pi$, and zero w.p. $1-\pi$. Gaussian 
noise is also considered with i.i.d. entries $\bV_t(m,n) \sim \calN(0,\sigma^2)$. Accordingly, the 
$t$-th acquired slice is formed as $\bY_t=\bOmega_t \odot (\bX_t+\bV_t)$. Fix $\sigma=10^{-3}$ and the
true rank $R=5$, while different values of $M,N,\hat{R},\pi$ are examined. Performance of 
Algorithm~\ref{tab:table_3} is tested for imputation of streaming tensor slices of relatively large size 
$M=N=10^3$, where a constant step size $(\bar{\mu}[t])^{-1}=10^{-2}$ 
is adopted. Various amounts of misses are examined, namely 
$1-\pi \in \{0.99,0.9,0.75\}$. Also, in accordance with the matrix completion setup 
select $\lambda=\sqrt{2MN\pi} \sigma$; 
see e.g.,~\cite{candes_moisy_mc}. Fig.~\ref{fig:fig_perf_synthdata_error} depicts the evolution of the 
estimation error $e_x[t]:=\|\bX_t-\hat{\bX}_t\|_F / \|\bX_t\|_F$, where it is naturally seen 
that as more data become available the tensor subspace is learned faster. It is also apparent 
that after collecting sufficient amounts of data the estimation error decreases 
geometrically, where finally the estimate $\hat{\bX}_t$ falls in the $\sigma^2$-neighborhood 
of the `ground truth' slice $\bX_t$. This observation suggests the linear convergence of 
Algorithm~\ref{tab:table_3}, and highlights the effectiveness of estimator (P3) in 
accurately reconstructing a large fraction of misses.

Here, Algorithm~\ref{tab:table_3} is also adopted to decompose large-scale, dense tensors 
and hence find the factors $\hat{\bA},\hat{\bB},\hat{\bC}$. For $T=10^4$ and for 
different slice sizes $M=N=10^2$ and $M=N=10^3$, the tensor may not even fit in main
memory to apply batch solvers naively. After running Algorithm \ref{tab:table_3} instead, 
Table~\ref{tab:table_5} reports the run-time under various 
amount of misses. One can see that smaller values of $\pi$ lead to shorter run-times since
one needs to carry out less computations per iteration~[c.f. $\mathcal{O}(|\Omega_t|\hat{R}^2)$]. Note 
that the MATLAB codes for these experiments are by no means optimized, so further reduction in 
run-time is possible with a more meticulous implementation. Another observation is that for 
decomposition of low-rank tensors, it might be beneficial from a computational complexity standpoint 
to keep only a small subset of entries. Note that if instead of employing a higher-order
decomposition one unfolds the tensor and resorts to the subspace tracking schemes developed 
in Section~\ref{sec:onlineMC} for the sake of imputation, each basis vector entails $10^6$ variables.
On the other hand, using tensor models each basis (rank-one) matrix entails only $2 \times 10^3$ variables. Once again, for comparison purposes there is no alternative online scheme that imputes the missing tensor entries, and offers a PARAFAC tensor decomposition after learning the tensor subspace~(see also Remark~\ref{rem:onthefly}).

%%%%%%%%%%%%%%%%%%%%%%%%%%%%%%%%%%%%%%%%%%%%%%%%%%%%%%%%%%%%%%%%%%%%%%%%%%%%%%%
%%%%%%%%%%%%%%%%%%%%%%%%%%%%%%%%%%%%%%%%%%%%%%%%%%%%%%%%%%%%%%%%%%%%%%%%%%%%%%%

%%%%%%%%%%%%%%%%%%%%%%%%%%%%%%%%%%%%%%%%%%%%%%%%%%%%%%%%%%%%%%%%%%%%%%%%%%%%%%%
\begin{table*}[t]
\caption{Tensor imputation. Run-time (seconds) for various sizes and amounts of misses when $R=10$, $T=10^4$.}
\label{tab:table_5}
\begin{center}
\begin{tabular} {|c|c|c|c|}
\hline
$M=N$ & $\pi=0.01$ & $\pi=0.1$ & $\pi=0.25$ \\
\hline
$10^2$ & $26$ & $132$ & $302$ \\
\hline
$10^3$ & $1.8 \times 10^3$ & $10^4$ & $3\times 10^4$ \\
\hline
\end{tabular}
\end{center}
\end{table*}
%%%%%%%%%%%%%%%%%%%%%%%%%%%%%%%%%%%%%%%%%%%%%%%%%%%%%%%%%%%%%%%%%%%%%%%%%%%%%%%

\subsection{Real tensor data tests}
Two real tensor data tests are carried out next, in the context of cardiac MRI 
and network traffic monitoring applications. 

\noindent\textbf{Cardiac MRI.} Cardiac MRI nowadays serves as a major imaging modality 
for noninvasive diagnosis of heart diseases in clinical practice~\cite{finn2006cardiac}. 
However, quality of MRI images is degraded as a result of fast acquisition 
process which is mainly due to patient's breath-holding time. This may render some 
image pixels inaccurate or missing, and thus the acquired image only consists of a 
subset of pixels of the high-resolution `ground truth' cardiac image. With this in mind, 
recovering the `ground truth' image amounts to imputing the missing pixels. Low-rank tensor
completion is well motivated by the low-intrinsic dimensionality of cardiac
MRI images~\cite{prism_haogao}. The FOURDIX dataset is considered for the ensuing tests,
and contains $263$ cardiac scans with $10$ steps of the entire cardiac cycle~\cite{osirix}.
Each scan is an image of size $512 \times 512$ pixels, which is divided into $64$ patches 
of $32 \times 32$ pixels. The $32 \times 32$ patches then form slices of the tensor 
$\underline{\bX} \in \mathbb{R}^{32\times 32\times 67,328}$. A large fraction ($75\%$ 
entries) of $\underline{\bX}$ is randomly discarded to simulate missing data. 

\begin{figure}[t]
\centering
\begin{tabular}{cc}
     \epsfig{file=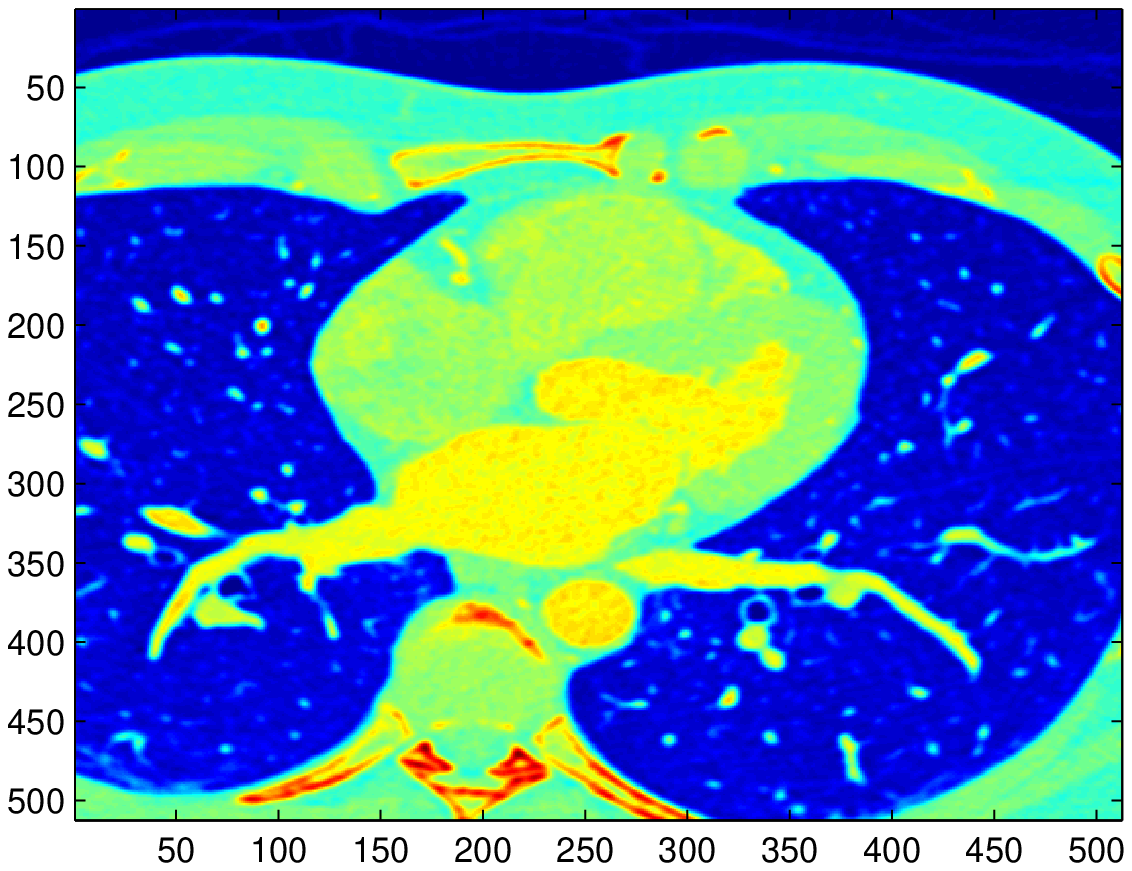,width=0.5
     \linewidth, height=2.3 in } &
     \epsfig{file=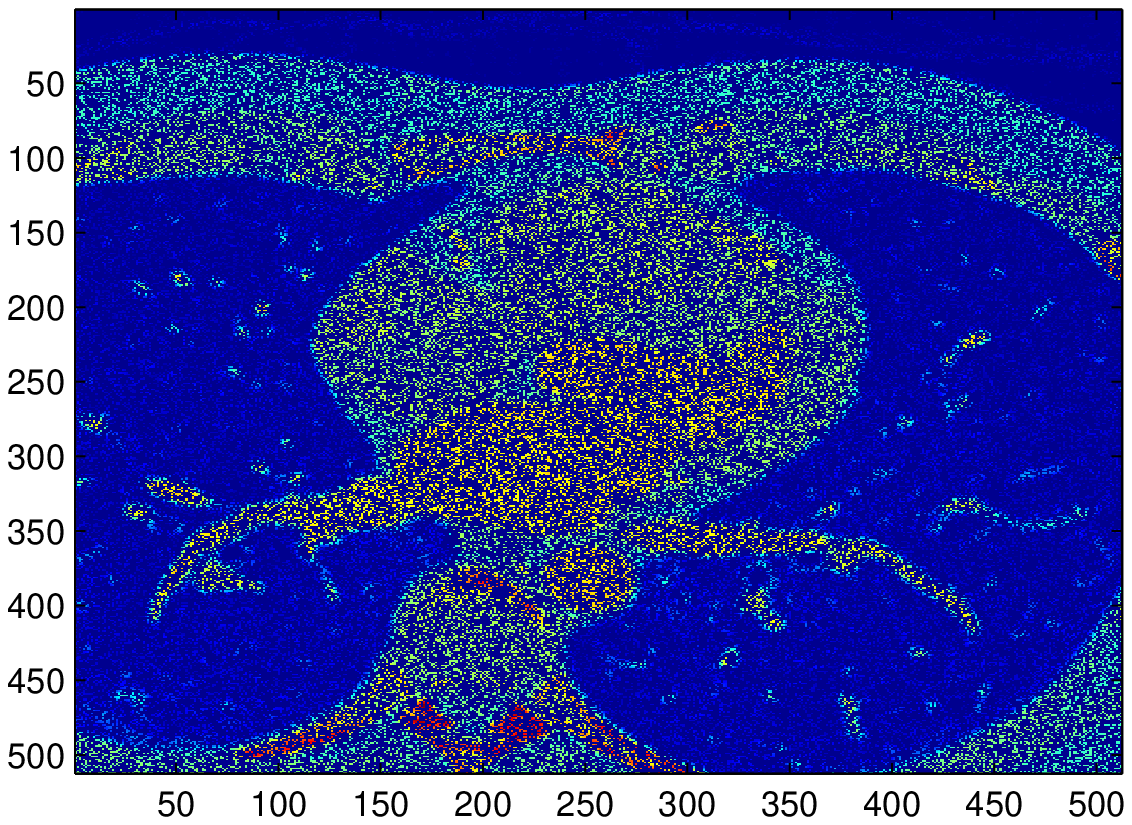,width=0.5
     \linewidth, height=2.3 in } \\
     (a) &
     (b) \\
     \epsfig{file=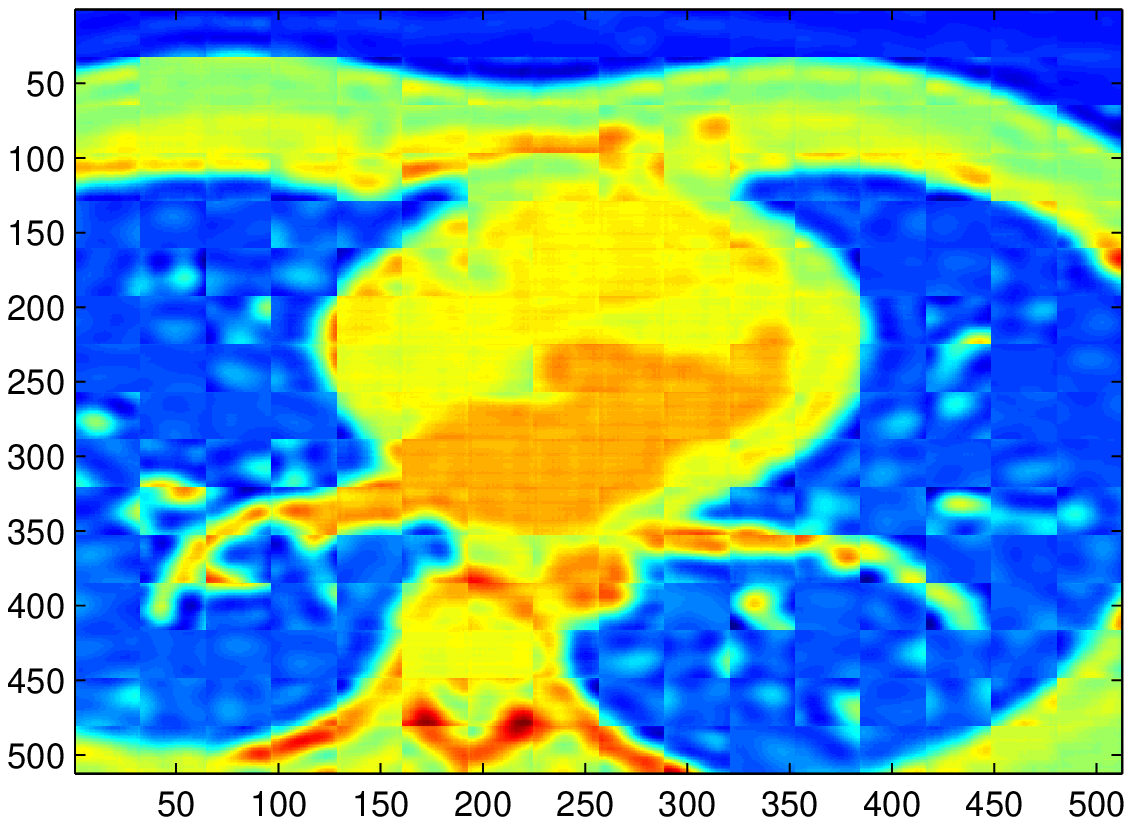,width=0.5
     \linewidth, height=2.3 in } &
     \epsfig{file=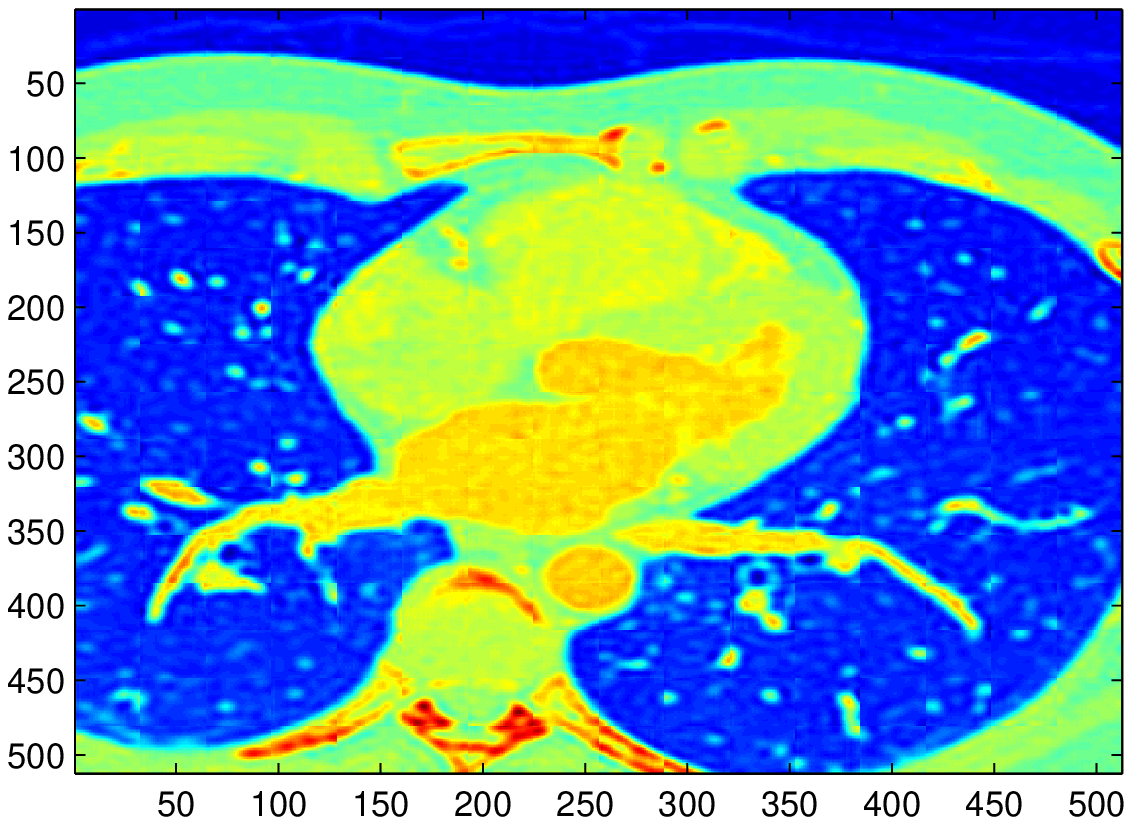,width=0.5
     \linewidth, height=2.3 in } \\
     (c) &
     (d) \\
  \end{tabular}
  \caption{Results of applying Algorithm~\ref{tab:table_3} to the cardiac MRI dataset FOURDIX~\cite{osirix}. (a)  Ground truth image, (b) acquired image with $75\%$ missing pixels, and the reconstructed image for rank (c) $\hat{R}=10$ and (d) $\hat{R}=50$. }
  \label{fig:fig_cardiac_mri}
\end{figure}

Imputing such a large, dense tensor via batch algorithms may be infeasible because
of memory limitations. The online Algorithm~\ref{tab:table_3} is 
however a viable alternative, which performs only $256 \hat{R}^2$ operations on average 
per time step, and requires storing only $256+64\hat{R}$ variables. For a candidate image, 
the imputation results of Algorithm~\ref{tab:table_3} are depicted in 
Fig.~\ref{fig:fig_cardiac_mri} for different choices of the rank $\hat{R}=10,50$. A 
constant step size $(\bar{\mu}[t])^{-1}=10^{-6}$ is chosen along with $\lambda=0.01$. 
Different choices of the rank $\hat{R}=10,50$ lead to $e_x=0.14,0.046$, respectively. Fig.~\ref{fig:fig_cardiac_mri}(a) 
shows the `ground truth' image, while Fig.~\ref{fig:fig_cardiac_mri}(b) depicts 
the acquired one with only $25\%$ available (missing entries are set to zero for display 
purposes.) Fig.~\ref{fig:fig_cardiac_mri}(c) also illustrates the reconstructed image 
after learning the tensor subspace for $\hat{R}=10$, and the result for $\hat{R}=50$ is shown in 
Fig.~\ref{fig:fig_cardiac_mri}(d). Note that although this test assumes misses 
in the spatial domain, it is more natural to consider misses in the frequency domain, 
where only a small subset of DFT coefficients are available. This model can be 
captured by the estimator (P5), by replacing the fidelity term with 
$\|\bOmega_{\tau} \odot \Psi (\bY_{\tau}-\bA {\rm diag}(\bxi_{\tau}) \bB')\|_F^2$, 
where $\Psi$ stands for the linear Fourier operator.

%%%%%%%%%%%%%%%%%%%%%%%%%%%%%%%%%%%%%%%%%%%%%%%%%%%%%%%%%%%%%%%%%%%%%%%%%%%%%%%
%%%%%%%%%%%%%%%%%%%%%%%%%%%%%%%%%%%%%%%%%%%%%%%%%%%%%%%%%%%%%%%%%%%%%%

\noindent\textbf{Tracking network-traffic anomalies.} In the backbone of large-scale 
IP networks, OD flows experience traffic volume anomalies due to e.g., equipment failures and 
cyberattacks which can congest the network~\cite{MC03}. Consider a network whose topology is
represented by a directed graph $G(\cN,\cL)$, where $\cL$ and $\cN$ denote the set of links and nodes of cardinality 
$|\cL|=L$ and $|\cN|=N$, respectively. Upon associating the weight $w_{i,j}>0,~(i,j)\in E$ with 
the $(i,j)$-th link, $G$ can be completely described by the weighted adjacency matrix 
$\bW \in \mathbb{R}^{N \times N}$. For instance $w_{i,j}$ can represent the link loads as will be shown 
later. In the considered network, a set of OD traffic flows $\cF$ with $|\cF|=F$ traverses the 
links connecting OD pairs. Let $r_{\ell,f} \in [0,1]$ denote the fraction of  $f$-th flow traffic 
at time $t$, say $x_{f,t}$, measured in e.g., packet counts, carried by link $\ell$. The overall 
traffic carried by link $\ell$ is then the superposition of the flow rates routed through link 
$\ell$, namely, $\sum_{f \in \cF} r_{\ell,f} x_{\ell,f}$. It is not uncommon for some of OD 
flows to experience anomalies. If $o_{f,t}$ denotes the unknown traffic volume anomaly of flow 
$f$ at time $t$, the measured link counts over link $\ell$ at time $t$ are then given by
\begin{align}
y_{\ell,t}=\sum_{f \in \cF} r_{\ell,f} (x_{f,t} + o_{f,t}) + v_{\ell,t},~\ell \in \cL \label{eq:y_l,t}
\end{align}
where $v_{\ell,t}$ accounts for measurement errors and unmodeled dynamics. In practice, missing link counts 
are common due to e.g., packet losses, and thus per time only a small fraction of links 
(indexed by $\Omega_t$) are measured. Note that only a small group of flows are anomalous, 
and the anomalies persist for short periods of time relative to the measurement 
horizon. This renders the anomaly vector $\bo_t=[o_{1,t},\ldots,o_{F,t}]^\prime \in \mathbb{R}^F$ sparse.  

In general, one can collect the partial link counts per time instant in a vector to form a 
(vector-valued) time-series, and subsequently apply the subspace tracking algorithms developed in e.g.,~\cite{jstsp_anomalography_2012} to unveil the anomalies in real time. %This approach 
%although benefits from well-developed matrix tools, it ignores knowledge of the network 
%topology. 
Instead, to fully exploit the data structure induced by the network topology, the link counts per time 
$t$ can be collected in an adjacency matrix $\bW_t$, with $[\bW_t]_{i,j}=y_{\ell,t}$ [edge $(i,j)$ 
corresponds to link $\ell$]. This matrix naturally constitutes the $t$-th slice of the tensor $\underline{\bY}$. 
Capitalizing on the spatiotemporal low-rank property of the nominal traffic as elaborated in 
Section~\ref{subsec:real_matrix_test}, to discern the anomalies a low-rank 
($\rho \ll R$) approximation of the incomplete tensor $\underline{\bOmega}\odot\underline{\bY}$ 
is obtained first in an online fashion using Algorithm~\ref{tab:table_3}. 
Anomalies are then unveiled from the residual of the approximation as elaborated next. 

Let $\{\bA[t],\bB[t]\}$ denote the factors of the low-dimensional tensor subspace learned 
at time $t$, and $\hat{\bY}_t=\bA[t] {\rm diag}(\bxi[t])\bB'[t] \in \mathbb{R}^{N \times N}$ 
the corresponding (imputed) low-rank approximation of the $t$-th slice. Form the residual 
matrix $\tilde{\bY}_t:= \bY_t-\hat{\bY}_t$, which is (approximately) zero in the absence of anomalies. 
Collect the nonzero entries of $\tilde{\bY}_t$ into the vector 
$\tilde{\by}_t \in \mathbb{R}^L$, and the routing variables $r_{\ell,f}$ [cf.~\eqref{eq:y_l,t}] 
into matrix $\bR \in \mathbb{R}^{L \times F}$. According to \eqref{eq:y_l,t}, one can postulate 
the linear regression model $\tilde{\by}_t=\bR \bo_t + \bv_t$ to estimate the sparse anomaly vector
$\bo_t \in \mathbb{R}^F$ from the imputed link counts. An estimate of $\bo_t$ can then be 
obtained via the least-absolute shrinkage and selection operator (LASSO)
\begin{align}
\nonumber \hat{\bo}_t:=\arg\min_{\bo \in \mathbb{R}^F} \|\tilde{\by}_t-\bR\bo\|^2 + \lambda_o \|\bo\|_1
\end{align}
where $\lambda_o$ controls the sparsity in $\hat{\bo}_t$ that is tantamount to the number of anomalies. 
In the absence of missing links counts,~\cite{xiaoli_ma_anomaly} has recently considered a batch tensor 
model of link traffic data and its Tucker decomposition to identify the anomalies. 

%%%%%%%%%%%%%%%%%%%%%%%%%%%%%%%%%%%%%%%%%%%%%%%%%%%%%%%%%%%%%%%%%%%%%%%%%%%%%%%
\begin{figure}[t]
\centering
\begin{tabular}{cc}
\hspace{-4mm}\epsfig{file=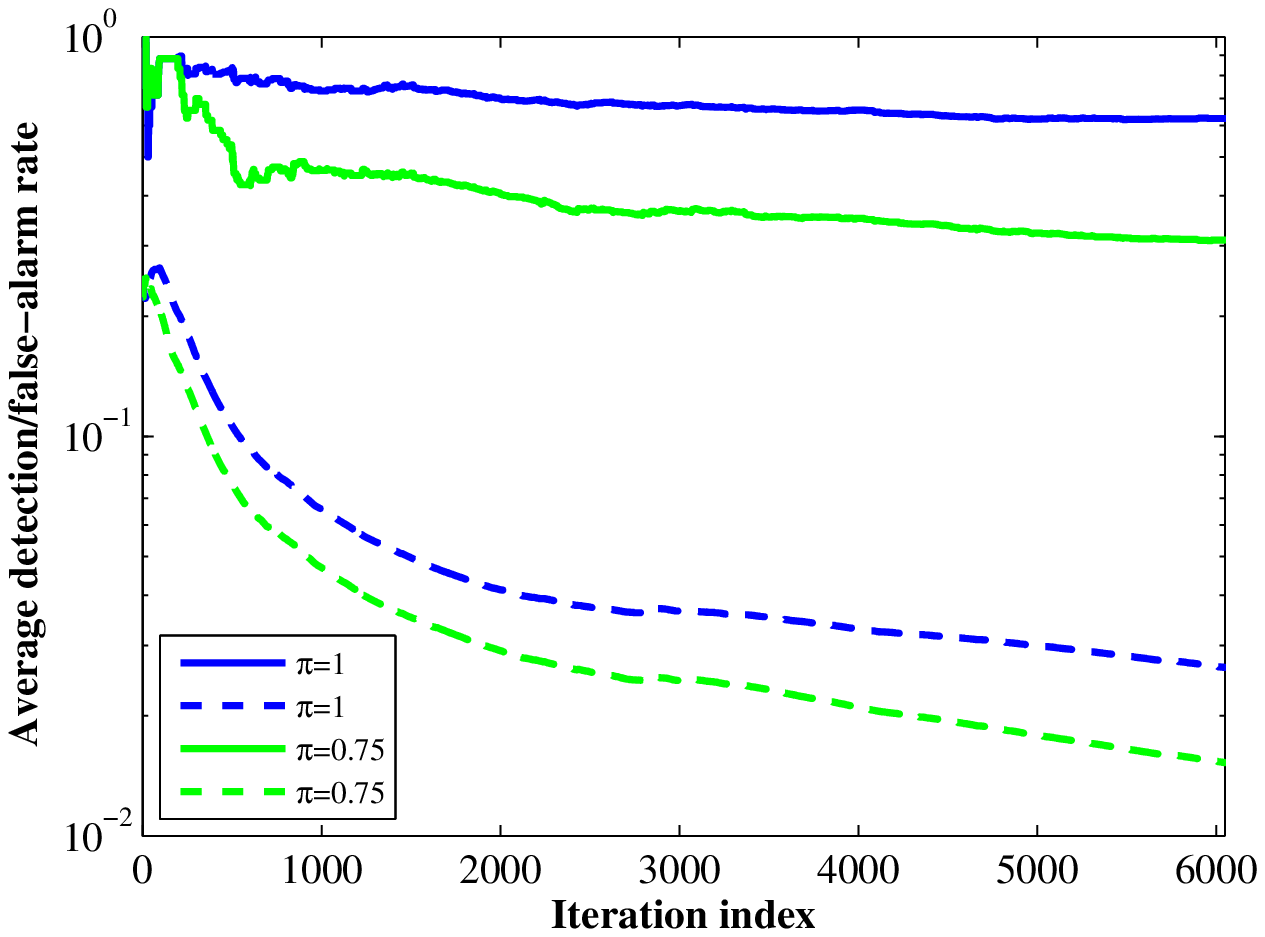,width=0.5
\linewidth, height=2.3 in } & \hspace{-8mm}
\epsfig{file=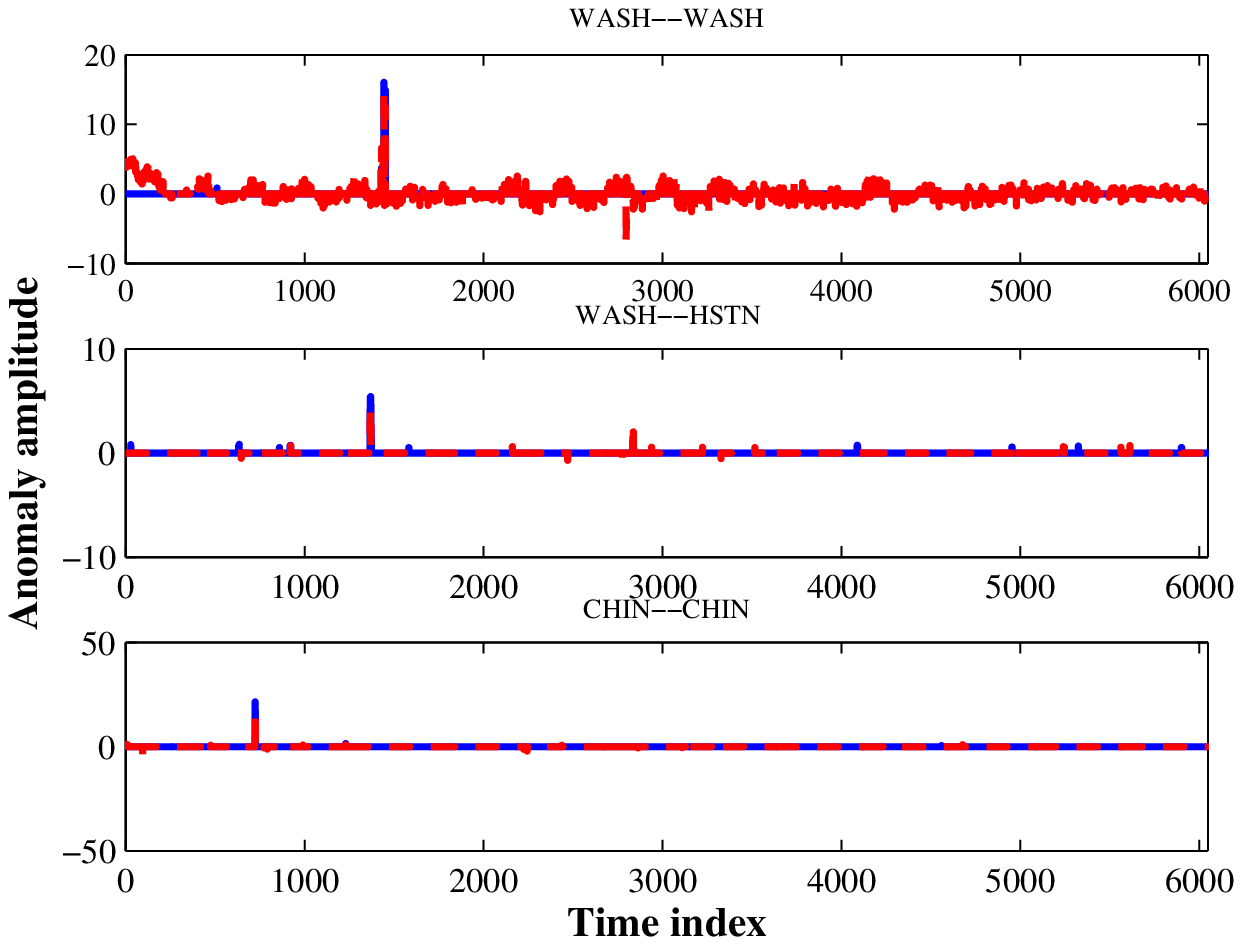,width=0.5
\linewidth, height=2.3 in } \\
(a) &
(b) \\
 \end{tabular}
 \vspace{-3mm}
 \caption{Tracking Internet-2 traffic anomalies for $\rho=18$. (a) Evolution of average detection (solid) and false-alarm (dashed) rates. (b) Estimated (red) versus true (blue) anomalies for three representative OD flows when $\pi=1$.}
 \label{fig:fig_perf_anomaly}
\end{figure}
%%%%%%%%%%%%%%%%%%%%%%%%%%%%%%%%%%%%%%%%%%%%%%%%%%%%%%%%%%%%%%%%%%%%%%%%%%%%%%%

The described tensor-based approach for network anomaly detection is tested on
the Internet-2 traffic dataset described in Section~\ref{subsec:real_matrix_test}, after fixing 
$\hat{R}=18$. Each tensor slice $\bY_t \in \mathbb{R}^{11 \times 11}$ contains only 
$41$ nonzero entries corresponding to the physical links. Define the sets 
$S_O[t]:=\{(i,j),~i\in [L],j \in [t]:~|o_j(i)| \geq \xi\}$ and $\bar{S}_O[t]:=\{(i,j),~i\in 
[L],j \in [t]:~|o_j(i)| \leq \xi\}$ for some prescribed threshold $\xi$. To evaluate the detection performance, 
the adopted figures of merit are the running-average detection and false-alarm rates 
$P_{\rm D}:=|S_O \cap S_{\hat{O}}|/|S_O|$ and $P_{\rm FA}:=|\bar{S}_O \cap S_{\hat{O}}|/|\bar{S}_O|$, 
respectively. Fig.~\ref{fig:fig_perf_anomaly}(a) depicts the time evolution of $P_{\rm D}$ 
and $P_{\rm FA}$ for $\pi=1$ (fully available data), and $\pi=0.75$. As 
more data becomes available, the traffic subspace is learned more accurately, and thus less 
false alarms are declared. For three representative OD flows, namely WASH-WASH, 
WASH--HSTN, and CHIN--CHIN, the true and estimated anomalies are depicted in 
Fig.~\ref{fig:fig_perf_anomaly}(b). One can see that the significant anomalies 
are correctly picked in real-time by the proposed estimator. Note that the online 
formulation (P5) can even accommodate slowly-varying network topologies in the tensor model, 
which is desirable for monitoring the `health state' of dynamic networks.

\section{Concluding Summary}
\label{sec:discussions}
This paper leverages recent advances in rank minimization for subspace tracking, 
and puts forth streaming algorithms for real-time, scalable decomposition of 
highly-incomplete multi-way Big Data arrays. 
For low-rank matrix data, a subspace tracker is 
developed based on an EWLS criterion regularized with the nuclear 
norm. Leveraging a separable characterization of nuclear-norm, both first- and second-order 
algorithms with complementary strengths are developed. In a stationary setting, 
the proposed algorithms asymptotically converge and provably offer the 
well-documented performance guarantees of the batch nuclear-norm regularized estimator. 
Under the same umbrella, an online algorithm is proposed for decomposing low-rank 
tensors with missing entries, which can accurately impute cardiac MRI images with 
up to $75\%$ missing entries.

There are intriguing unanswered questions beyond the scope of this paper, but worth 
pursuing as future research. One such question pertains to the convergence analysis of the 
accelerated SGD algorithm either by following the adopted proof methodology, or, e.g., the 
alternative techniques used in~\cite{kostas_icassp2014_cnvg}.
Real-time incorporation of the spatiotemporal correlation between 
the unknowns by means of kernels or suitable statistical models is another 
important avenue to explore. Also, relaxing the 
qualification constraint for optimality is important for real-time applications
in dynamic environments, 
where the learned subspace could conceivably change with time.

\appendix

% % % % % % % % % % % % % % % % % % % % % % % % % % % % % % % % % % % % % % % %
%                         Appendix A                                          %
% % % % % % % % % % % % % % % % % % % % % % % % % % % % % % % % % % % % % % % %

\noindent\normalsize \emph{\textbf{Proof of Proposition~\ref{prop:prop_3}}.}
For the subspace sequence $\{\bL[t]\}$ suppose that $\lim_{t \rightarrow \infty} \nabla C_t(\bL[t])=\mathbf{0}$. Then, due to the uniqueness of $\bq[t]=\arg\min_{\bq} g_t(\bL[t],\bq)$, Danskin's Theorem~\cite{Bers} implies that
\begin{align}
\lim_{t \rightarrow \infty} \frac{1}{t} \left( \cP_{\Omega_t} (\bY_t-\bL[t]\bQ'[t])\bQ[t] - \lambda \bL[t] \right) = \mathbf{0} \label{eq:lim_grad_C_L}
\end{align}
holds true almost surely, where $\bQ[t] \in \mathbb{R}^{P \times t}$ satisfies
\begin{align}
\bL'[t] \cP_{\Omega_t}(\bY_t-\bL[t]\bQ'[t]) - \lambda \bQ'[t] = \mathbf{0}. \label{eq:grad_C_Q}
\end{align}
Consider now a subsequence $\{\bL[t_k],\bQ[t_k]\}_{k=1}^{\infty}$ which satisfies~\eqref{eq:lim_grad_C_L}--\eqref{eq:grad_C_Q} as well as the qualification constraint $\|\cP_{\Omega_{t_k}}(\bY_{t_k}-\bL[t_k]\bQ'[t_k])\| \leq \lambda$. The rest of the proof then verifies that $\{\bL[t_k],\bQ[t_k]\}$ asymptotically fulfills the optimality conditions for (P1). To begin with, the following equivalent formulation of (P1) is considered at time $t_k$, which no longer involves the non-smooth nuclear norm. 
\begin{align}
\text{(P5)}~~~\min_{\substack{\{\bX \in \mathbb{R}^{P \times P},\bW_1 \in \mathbb{R}^{P \times t_k} \\ \bW_2 \in \mathbb{R}^{t_k \times t_k}\}}}& \left[\frac{1}{2t_k}\|\cP_{\Omega_{t_k}}(\bY_{t_k} - \bX)\|_{F}^{2} + \frac{\lambda}{2t_k}\left\{\tr(\bW_1)+\tr(\bW_2)\right\} \right]\nonumber\\
\text{s. to}\quad & \bW:=\left(\begin{array}{cc}\bW_1&\bX\\
\bX'& \bW_2\end{array}\right)  \succeq \mathbf{0} \nonumber
\end{align}
To explore the optimality conditions for (P5), first form the Lagrangian 
\begin{align}
\cL_{t_k}(\bX,\bW_1,\bW_2;\bM)=\frac{1}{2t_k}\|\cP_{\Omega_{t_k}}(\bY_{t_k} - \bX)\|_{F}^{2} +
\frac{\lambda}{2t_k}\left(\tr\{\bW_1\}+\tr\{\bW_2\}\right)-\langle \bM,\bW \rangle.
\end{align}
where $\bM$ denotes the dual variables associated with the positive semi-definiteness constraint in (P5). For notational convenience, partition $\bM$ into four blocks, namely $\bM_1:=[\bM]_{11}$,
$\bM_2:=[\bM]_{12}$, $\bM_3:=[\bM]_{22}$, and $\bM_4:=[\bM]_{21}$, in accordance with the block
structure of $\bW$ in~(P5), where $\bM_1$ and $\bM_3$ are $P \times P$ and $t_k\times t_k$ matrices.
The optimal solution to (P1) must: (i) null the (sub)gradients
\begin{align}
&\nabla_{\bX}{\cL}_{t_k}(\bX,\bW_1,\bW_2;\bM)=-\frac{1}{t_k}\cP_{\Omega_{t_k}}(\bY_{t_k}-\bX) - \bM_2 - \bM'_4 \label{eq:dlx}\\
&\nabla_{\bW_1}{\cL}_{t_k}(\bX,\bW_1,\bW_2;\bM)=\frac{\lambda}{2t_k}\bI_L - \bM_1 \label{eq:dlw_1}\\
&\nabla_{\bW_2}{\cL}_{t_k}(\bX,\bW_1,\bW_2;\bM)=\frac{\lambda}{2t_k}\bI_{t_k} - \bM_3 \label{eq:dlw_2}
\end{align}
(ii) satisfy the complementary slackness condition $\langle \bM,\bW \rangle=0$;
(iii) primal feasibility $\bW\succeq \mathbf{0}$; and (iv) dual feasibility $\bM \succeq \mathbf{0}$.

Introduce the candidate primal variables $\bX[k]:=\bL[t_k]\bQ'[t_k]$, $\bW_1[k]:=\bL[t_k]\bL'[t_k]$ and $\bW_2[k]:=\bQ[t_k]\bQ'[t_k]$;
and the dual variables $\bM_1[k]:=\frac{\lambda}{2t_k}\bI_L$,
$\bM_3[k]:=\frac{\lambda}{2t_k}\bI_t$, $\bM_2[k]:=-(1/2t_k)\cP_{\Omega_{t_k}}(\bY_{t_k}-\bL[t_k]\bQ'[t_k])$, and $\bM_4[k]:=\bM_2'[k]$. Then, it can be readily verified that (i), (iii) and (iv) hold. Moreover, (ii) holds since
\begin{align}
\langle \bM[k],\bW[k] \rangle &= \langle \bM_1[k],\bW_1[k]\rangle + \langle \bM_2[k],
\bX[k] \rangle + \langle \bM_2'[k],\bX'[k]\rangle + \langle \bM_3[k],\bW_2[k] \rangle \nonumber\\
&=\frac{\lambda}{2t_k} \langle \bI_L, \bL[t_k]\bL'[t_k]   \rangle + \frac{\lambda}{2t_k} \langle \bI_{t_k}, \bQ[t_k]\bQ'[t_k]\rangle - \frac{1}{t_k} \langle \cP_{\Omega_{t_k}}(\bY_{t_k}-\bL[t_k]\bQ'[t_k]), \bL[t_k]\bQ'[t_k] \rangle \nonumber\\
&=\frac{1}{2t_k} \langle \bL[t_k], \lambda\bL[t_k] - \cP_{\Omega_{t_k}}(\bY_{t_k}-\bL[t_k]\bQ'[t_k])\bQ[t_k] \rangle \nonumber\\ 
&\hspace{40mm} + \frac{1}{2t_k} \langle \bQ'[t_k], \lambda \bQ'[t_k] - \bL'[t_k]\cP_{\Omega_{t_k}}(\bY_{t_k}-\bL[t_k]\bQ'[t_k]) \rangle \nonumber\\
&=\frac{1}{2t_k} \langle \bL[t_k], \lambda \bL[t_k] - \cP_{\Omega_{t_k}}(\bY_{t_k}-\bL[t_k]\bQ'[t_k])\bQ[t_k] \rangle \nonumber
\end{align}
where the last equality is due to \eqref{eq:grad_C_Q}. Putting pieces together, the Cauchy-Schwartz inequality implies that
\begin{align}
&\lim_{k \rightarrow \infty} |\langle \bM[k],\bW[k] \rangle| \leq  \sup_{k} \|\bL[t_k]\|_F \nonumber\\& 
\hspace{2cm}\times \lim_{k \rightarrow \infty} \|\frac{1}{2t_k} \left( \lambda \bL[t_k] - \cP_{\Omega_{t_k}}(\bY_{t_k}-\bL[t_k]\bQ'[t_k])\bQ[t_k] \right)\|_F \nonumber\ = 0
\end{align}
holds almost surely due to~\eqref{eq:lim_grad_C_L}, and (A3) which says $\|\bL[t_k]\|_F$ is bounded. All in all, $\lim_{k \rightarrow \infty} \langle \bM[k],\bW[k] \rangle = 0$, which completes the proof.\hfill$\blacksquare$

% % % % % % % % % % % % % % % % % % % % % % % % % % % % % % % % % % % % % % % %
%                         References                                          %
% % % % % % % % % % % % % % % % % % % % % % % % % % % % % % % % % % % % % % % %

%\newpage
\bibliographystyle{IEEEtranS}
\bibliography{IEEEabrv,biblio}

\newpage

\end{document}